%% file: arxiv.tex
\documentclass[11pt]{article}
\PassOptionsToPackage{hyperfootnotes=false}{hyperref}

\usepackage[preprint]{acl}

\usepackage{times}
\usepackage{latexsym}

\usepackage{amsmath}
\usepackage{amssymb}
\usepackage{graphicx}
\usepackage{booktabs}
\usepackage{algorithm}
\usepackage{algpseudocode}
\usepackage{hyperref}
\usepackage{url}
\usepackage{multirow}
\usepackage{diagbox}
\usepackage{colortbl}
\usepackage{tabularx}
\usepackage{subfigure}
\usepackage{pifont}
\usepackage{color, xcolor}
\usepackage{soul}
\definecolor{red0}{RGB}{255,192,203}
\definecolor{green0}{RGB}{127,255,212}
\definecolor{AccessOutput}{HTML}{F5F7F9}
\definecolor{AccessProb}{HTML}{F8F8F6}
\definecolor{AccessWhite}{HTML}{F6F5F8}
\colorlet{AccessInternal}{AccessWhite}
\definecolor{LSCMethod}{HTML}{E7F2FA}
\definecolor{LSCAccent}{HTML}{3E78A0}
\definecolor{LSCSpine}{HTML}{707070}
\definecolor{BestScore}{HTML}{B22222}
\definecolor{SecondScore}{HTML}{1F5AA6}
\newcommand{\lscmethodrow}{\rowcolor{LSCMethod}}
\newcommand{\bestscore}[1]{\textcolor{BestScore}{\textbf{#1}}}
\newcommand{\secondscore}[1]{\textcolor{SecondScore}{\underline{#1}}}
\newcommand{\scorelegend}{Best and second-best results in each setting are shown in \textcolor{BestScore}{red}/\textbf{bold} and \textcolor{SecondScore}{blue}/\underline{underlined}.}
\defcitealias{kuhn2023semantic}{Kuhn et al.}
\defcitealias{tong2025sre}{Tong et al.}
\defcitealias{ren2022out}{Ren et al.}
\defcitealias{manakul2023selfcheckgpt}{Manakul et al.}
\defcitealias{liu2020energy}{Liu et al.}
\defcitealias{malinin2020uncertainty}{Malinin and Gales}
\defcitealias{farquhar2024semantic}{Farquhar et al.}
\defcitealias{nikitin2024kernel}{Nikitin et al.}
\defcitealias{li2025semanticgraph}{Li et al.}
\defcitealias{kunitomo2026evidential}{Kunitomo-Jacquin et al.}
\defcitealias{chen2024inside}{Chen et al.}
\defcitealias{liu2025attention}{Liu et al.}
\defcitealias{bakman2024mars}{Bakman et al.}
\defcitealias{aichberger2025sdlg}{Aichberger et al.}
\newcommand{\namecite}[1]{[\citetalias{#1}]}

\usepackage[T1]{fontenc}

\usepackage[utf8]{inputenc}

\usepackage{microtype}

\usepackage{inconsolata}

\usepackage{enumitem}

\title{Lowest Span Confidence: Zero-Shot Hallucination Detection from a Single LLM Response}

\author{
\textbf{Yitong Qiao\textsuperscript{1,2,}\thanks{Work done during Yitong's internship at Ant Group.}},
\textbf{Licheng Pan\textsuperscript{1}},
\textbf{Yu Mi\textsuperscript{1}},
\textbf{Lei Liu\textsuperscript{1,2}},
\textbf{Yue Shen\textsuperscript{2}}\\
\textbf{Jian Wang\textsuperscript{2}},
\textbf{Jinjie Gu\textsuperscript{2}},
\textbf{Fei Sun\textsuperscript{3}},
\textbf{Zhixuan Chu\textsuperscript{1}}\\[0.5em]
\textsuperscript{1}Zhejiang University\quad
\textsuperscript{2}Ant Healthcare, Ant Group\\
\textsuperscript{3}Institute of Computing Technology, Chinese Academy of Sciences\\[0.3em]
\small{\textbf{Correspondence:} \href{mailto:zhixuanchu@zju.edu.cn}{zhixuanchu@zju.edu.cn}}
}

\hypersetup{
  pdfauthor={Yitong Qiao, Licheng Pan, Yu Mi, Lei Liu, Yue Shen, Jian Wang, Jinjie Gu, Fei Sun, Zhixuan Chu}
}

\begin{document}
\maketitle

\begin{abstract}
Hallucinations in Large Language Models (LLMs), \textit{i.e.}, plausible but non-factual generations, pose a significant challenge to reliable deployment in high-stakes environments. However, many existing hallucination detectors require expensive repeated sampling for consistency checks or access to model-internal states unavailable in common API-based scenarios. To this end, we propose an efficient zero-shot metric called \textbf{Lowest Span Confidence (LSC)} for hallucination detection under minimal resource assumptions. Concretely, LSC evaluates the local confidence of adjacent complete-word spans. By selecting the lowest aggregated confidence across neighboring words whose token widths can vary, LSC captures localized uncertainty associated with factual inconsistency. This boundary-aligned smoothing reduces the global dilution of perplexity and the sensitivity of minimum token probability to isolated noise. Our main evaluation spans four model families \{Llama-2, Qwen2.5, Gemma-2, Mistral\} and seven benchmarks \{NQ, TriviaQA, SQuAD, CoQA, HotpotQA, RAGTruth, FELM\}. Additional analyses examine word reconstruction, span width, and the role of adjacency in preserving local confidence. Across these settings, LSC is competitive with methods that use multiple responses or model-internal information while requiring \textbf{only one response and its output token probabilities}, without training a separate detector or using an auxiliary model.
\end{abstract}

\section{Introduction}

Large Language Models (LLMs) have achieved remarkable performance across diverse natural language processing tasks, enabling their deployment in high-stakes domains such as healthcare \cite{wang2025trustworthy}, finance \cite{dong2025large}, and autonomous agents \cite{wang2024survey}. However, their tendency to generate fluent yet factually incorrect content—commonly known as hallucination—remains a critical barrier to reliable deployment \cite{he2024llm, manakul2023selfcheckgpt, li2024dawn}. In contexts where factual accuracy is paramount, even a single hallucinated claim can erode user trust and lead to harmful outcomes, motivating practical hallucination detection.

Existing hallucination detectors make different trade-offs between generation cost, model access, and the resolution of their uncertainty signal. \emph{Comparing multiple responses} measures semantic or lexical agreement among several outputs for the same input \cite{manakul2023selfcheckgpt, zhang2023sac3, chen2024inside}. Variation across responses can reveal uncertainty, but repeated decoding increases latency. Agreement measures generation stability rather than factual correctness, so a confidently repeated error may go undetected.

\emph{Using model-internal information} relies on white-box signals such as attention weights or hidden states \cite{he2024llm, chuang2024lookback}. These representations can support reference-free detection, but most commercial LLM APIs do not expose them and instead return generated tokens with, at most, output log-probabilities.

\emph{Using one-response probabilities} avoids both repeated generation and internal activations. Yet existing scores discard complementary structure. Perplexity averages uncertainty over the whole response and can dilute a short factual error, whereas Min-P keeps one extreme token and is sensitive to rare but correct words or unusual token splits.

In this work, we propose Lowest Span Confidence (LSC), a simple yet effective zero-shot metric for hallucination detection that operates under minimal resource assumptions. Unlike global metrics such as perplexity, whose signal can be diluted by long, high-confidence contexts, or single-token approaches like minimum token probability (Min-P), which are highly sensitive to local noise, LSC aggregates the model’s output probabilities over adjacent complete-word spans. By selecting the span with the lowest aggregated confidence across neighboring words whose token widths can vary, LSC targets localized uncertainty associated with factual inconsistencies. LSC requires one response and its token probabilities, making it compatible with APIs that expose these probabilities.

Our main contributions are:

\begin{itemize}[leftmargin=*]
\item We identify a practical gap in hallucination detection. Response consistency and model internals require repeated generation or white-box access, while one-response probability scores overlook local span structure. We propose \textbf{Lowest Span Confidence (LSC)}, a zero-shot metric using one response and its output token probabilities.
\item We reconstruct words from model tokens and average confidence over adjacent word pairs, bridging global averaging and token-level sensitivity. This smoothing retains continuous low-confidence regions while reducing isolated noise.
\item We conduct experiments with four model families across seven benchmarks. LSC achieves competitive performance against approaches using multiple generations or hidden states.
\end{itemize}

\section{Related Work}
\label{sec:related_work}

Table \ref{tab:access_taxonomy} groups detectors by their response budget and access to selected-token probabilities, full next-token distributions, or model internals.

\input{tables/arxiv/access_taxonomy}

\paragraph{Using Output Probabilities.}
Probability-based methods assign uncertainty scores from the model's output distribution. PPL and generation NLL average token log-probabilities, whereas Min-P retains the least probable token. These basic scoring rules are commonly used in uncertainty studies \cite{malinin2020uncertainty,ren2022out,manakul2023selfcheckgpt}. MARS \cite{bakman2024mars} reweights token probabilities using a learned semantic-importance model. LSC instead averages confidence within complete words and adjacent word pairs without an auxiliary model.

\paragraph{Comparing Multiple Responses.}
SelfCheckGPT \cite{manakul2023selfcheckgpt} asks the model for several responses and checks whether they support the original answer. SAC3 \cite{zhang2023sac3} compares responses from different models or from rephrased questions. EigenScore \cite{chen2024inside} and Lexical Similarity \cite{kuhn2023semantic} measure similarity in representation space or wording. AGSER \cite{liu2025attention} generates answers from attention-selected subsets of the query. SDLG \cite{aichberger2025sdlg} steers the model toward semantically diverse alternatives and uses them to estimate semantic uncertainty. Repeated generation increases inference cost, and response agreement alone need not reveal a confidently repeated error \cite{zhang2023sac3, chen2024inside}.

\paragraph{Using Model Internals.}
Methods train classifiers on hidden states \cite{azaria2023internal,he2024llm,orgad2024llms} or use attention values \cite{chuang2024lookback}. These approaches may require labeled data, training, or internal representations unavailable through hosted APIs.

\paragraph{Combining Local and Global Confidence.}
ConfFuse \cite{zhang2026conffuse} combines learned chain-level confidence with step-level scores from a process reward model for mathematical reasoning. In speech recognition, \citet{corpataux2026whisper} use local confidence contrasts and contextual features to train a hallucination classifier. These studies also use local signals, but target different tasks and rely on trained predictors. LSC directly scores adjacent word pairs without training a detector.

\section{Methodology}
\label{sec:methodology}
We consider detection from one response and its selected-token probabilities, without repeated generation or model-internal states. Lowest Span Confidence (LSC) finds the least confident local span while reducing sensitivity to unusual tokens.

\subsection{Problem Setup and Design Principle}
\label{sec:problem_formulation}

Let $x$ denote a user query and $\mathcal{M}$ a large language model. The model generates a response $y=\{t_1,t_2,\ldots,t_T\}$. For each generated token, it returns the conditional probability
$p_i=P(t_i\mid x,y_{<i})$.
We compute a score $R_{\mathrm{LSC}}(y)$, where larger values indicate greater factual-error risk, and identify its source span. LSC uses neither reference answers nor any external fact checker.

\paragraph{Motivation 1: Whole-response confidence can hide a short factual error.}
Probability-based detectors often summarize a response in one of two ways. Perplexity averages uncertainty over the entire response, so a short low-confidence phrase can be hidden by the remaining high-confidence text. Min-P uses only the least confident token, so it may flag a rare but correct name or an unusual token split. We seek a local signal that retains confidence drops without depending on a single token.

\paragraph{Motivation 2: Fixed token windows do not cover the same text.}
A fixed number of model tokens can cover different amounts of text. Depending on the tokenizer, two tokens may form part of one word, one complete word, or two short words. We rebuild complete words from token pieces and compute one confidence value for each word. Scoring neighboring words then keeps a common text unit despite differences in model tokenization.

\subsection{Lowest Span Confidence}
\label{sec:method_lsc}

LSC examines every pair of neighboring words and keeps the least confident pair. Each pair contains two complete words with a variable number of model tokens, avoiding a fixed token-window size. Algorithm~\ref{alg:lsc} summarizes the procedure.

\begin{algorithm}[!t]
\caption{Lowest Span Confidence.}
\label{alg:lsc}
\small
\begin{algorithmic}[1]
\Require Token pieces $z_{1:T}$ of a response containing non-whitespace text and selected log-probabilities $\ell_{1:T}$
\Ensure Risk $R$ and token span $I^\star$
\State Set $p_i\gets\exp(\ell_i)$ and reconstruct word groups $G_{1:M}$
\For{$k=1,\ldots,M$}
    \State $q_k\gets |G_k|^{-1}\sum_{i\in G_k}p_i$
\EndFor
\If{$M=1$}
    \State \Return $-\log q_1,G_1$
\EndIf
\State $k^\star\gets\arg\min_{1\le k<M}(q_k+q_{k+1})/2$
\State $R\gets-\log\big((q_{k^\star}+q_{k^\star+1})/2\big)$
\State $I^\star\gets G_{k^\star}\cup G_{k^\star+1}$
\State \Return $R,I^\star$
\end{algorithmic}
\end{algorithm}

We rebuild $M$ surface words from decoded token pieces, defining each word as a maximal non-whitespace interval. Attached punctuation and clitics stay within the word. Let $G_k$ span the first through last token overlapping word $k$. Appendix \ref{app:word_reconstruction} gives the decoding and overlap rules. We compute its mean selected-token confidence as
\begin{equation}
    q_k=\frac{1}{|G_k|}\sum_{i\in G_k}p_i.
\end{equation}
LSC gives both words equal weight in each pair:
\begin{equation}
\begin{aligned}
    C^{\mathrm{word}}_{k,2}&=\frac{q_k+q_{k+1}}{2},\\
    C_{\mathrm{LSC}}(y)&=\min_{1\le k<M}C^{\mathrm{word}}_{k,2},\\
    R_{\mathrm{LSC}}(y)&=-\log C_{\mathrm{LSC}}(y).
\end{aligned}
\end{equation}
When $M=1$, we use $q_1$. The first average prevents extra weight for words with more token pieces. The second reduces the influence of an isolated low-confidence word while retaining contiguous confidence drops. The pair mean also equals expected token confidence when choosing one word uniformly from the pair, then one token uniformly within that word. This is a local confidence statistic, not the probability of a word or sequence.

\paragraph{Why adjacent words?}
Two words form the smallest window that combines neighboring-word confidence. This is a local smoothing scale, not an assumed factual-claim length. Every adjacent pair is scored, so a long claim can span several windows. Arithmetic averaging retains the neighbor's contribution even when one confidence approaches zero, unlike geometric or harmonic averaging. Section \ref{sec:analysis} and Appendix \ref{app:design_ablations} compare means, span widths, adjacency, and sentence-level scoring.

\section{Experiments}
\input{tables/main_results}
\subsection{Experimental Setup}

\paragraph{Datasets.}
We use four standard QA benchmarks: Natural Questions (NQ) \cite{kwiatkowski2019natural}, TriviaQA \cite{joshi2017triviaqa}, the answerable portion of SQuAD~2.0 \cite{rajpurkar2018squad}, and CoQA with its full dialogue history \cite{reddy2019coqa}. We select 2,000 examples per model and dataset, with method-specific scoring counts given in the appendices. We also use 2,000 multi-hop questions from HotpotQA \cite{yang2018hotpotqa}. Beyond generated QA answers, RAGTruth \cite{niu2024ragtruth} provides 899 released responses with human labels for responses and hallucinated spans. FELM \cite{chen2023felm} provides 847 open-ended responses with segment-level factuality labels in five domains: world knowledge, science, writing and recommendation, reasoning, and mathematics.

\paragraph{Models.}
We select one similarly sized instruction-tuned model from each of four open-source families: \texttt{Llama-2-7B-Chat} \cite{touvron2023llama}, \texttt{Qwen2.5-7B-Instruct} \cite{qwen2024qwen25}, \texttt{Gemma-2-9B-IT} \cite{gemmateam2024gemma2}, and \texttt{Mistral-7B-Instruct-v0.3} \cite{jiang2023mistral}. Together, they provide a controlled comparison across four model families at roughly 7B--9B parameters. All models use their official checkpoints without additional fine-tuning.

\paragraph{Baselines.}
Table \ref{tab:access_taxonomy} groups methods by response budget and model access. Among single-response methods, PPL, length-normalized generation NLL (G-NLL), Min-P, and LSC use selected-token probabilities $p(y_t)$. Energy \cite{liu2020energy} and Mean Token Entropy \cite{malinin2020uncertainty} require the full next-token distribution $p(\cdot)$. Multi-response methods include Lexical Similarity \cite{kuhn2023semantic}, SRE \cite{tong2025sre}, LN-Entropy \cite{malinin2020uncertainty}, Semantic Entropy \cite{farquhar2024semantic}, KLE \cite{nikitin2024kernel}, SGD \cite{li2025semanticgraph}, and EVSE \cite{kunitomo2026evidential}. EigenScore \cite{chen2024inside} and AGSER \cite{liu2025attention} additionally require internal model states. We evaluate AGSER and SRE in all 16 model--dataset settings. SRE may fail to obtain valid reformulations, so we report the fraction of responses it scores. We also compare with Confidence + MARS using the authors' released importance predictor. SDLG appears in the access taxonomy as related work. Appendix \ref{app:more_baseline} describes the methods, including the MARS comparison.

\paragraph{Evaluation Metrics.}
We report AUROC, where a higher value means that a method better separates correct and incorrect responses. We separate the datasets according to how these labels are obtained. For NQ, TriviaQA, SQuAD, CoQA, and HotpotQA, we compare each response with a reference answer. RAGTruth and FELM instead provide human factuality labels. A FELM response is labeled as hallucinated when at least one annotated segment is factually incorrect. We do not average reference-based and human-label results.

\paragraph{Correctness Measure.}
For NQ, TriviaQA, SQuAD, and CoQA, AUC$_r$ treats a response as correct when its highest ROUGE-L F1 score against the reference answers \cite{lin2004rouge} is at least 0.5. AUC$_s$ treats it as correct when its highest sentence-embedding similarity to a reference is at least 0.9. HotpotQA uses token F1 with a threshold of 0.5. Exact match would be too strict because a correct paraphrase need not repeat the reference wording. References are evaluation labels, not detector inputs. Appendix \ref{app:evaluation_protocols} specifies the procedure.

\paragraph{Implementation Details.}
We generate the main QA response with greedy decoding. Baselines that share additional responses use six samples generated with temperature 1.0, top-$p$ 0.9, and top-$k$ 50. LSC uses the main response's token probabilities and pieces, with the same word-pair definition in every main experiment. For FELM, each scoring model reads the same prompt and released response with teacher forcing. This tests transfer across scoring models, not the quality of their generations. We normalize ordinary sequence likelihood scores by the number of valid output tokens. PPL is the exponential of this mean NLL. MARS uses its own importance-weighted rule, described in Appendix \ref{app:mars_comparison}. Appendix \ref{app:runtime} gives the runtime measurements.

\subsection{Main Results}

\paragraph{Reference-based QA Correctness.}
Table \ref{tab:main_results} reports AUC$_s$/AUC$_r$ for the same greedy responses in each model and dataset. LSC improves the basic one-response probability scores in most comparisons. It also remains competitive with methods using complete next-token distributions, multiple responses, or internal model states. In the additional comparison with Confidence + MARS, LSC gains 15.76/12.72 AUC$_s$/AUC$_r$ points on average across the 16 QA settings (Appendix \ref{app:mars_comparison}).

\paragraph{Human-annotated Factuality.}
Table \ref{tab:ragtruth_response} evaluates the released RAGTruth responses using human hallucination labels. On all 899 responses, LSC obtains 58.74 AUROC, compared with 47.32 for PPL and 42.82 for Min-P. It also improves the result for each Llama-2 scorer separately, with its highest task-level AUROC on QA. Appendix \ref{app:ragtruth} reports task and error-type results, together with overlap between detected and human-marked spans.
\input{tables/broader_results}

\paragraph{Open-ended Factuality.}
Table \ref{tab:felm_results} reports an equal-weight macro average over the five FELM domains. LSC reaches 59.55 AUROC and leads the selected-token baselines in this table for all three scoring models. Methods with complete next-token access are reported separately. Appendix \ref{app:felm} gives pooled AUROC and AUPRC, alongside the full domain-level scoring and evaluation protocol.
\input{tables/felm_results}

\paragraph{Multi-hop QA.}
Table \ref{tab:hotpot_results} reports HotpotQA results, where questions can require several pieces of supporting information. Among the selected-token baselines, LSC is best for all three model families, reaching 65.66, 79.92, and 69.80 AUROC.
\input{tables/hotpot_results}

\subsection{Why Does LSC Work?}
\label{sec:analysis}
\input{tables/arxiv/mechanism_evidence}

\paragraph{Aggregation Function.}
Arithmetic, geometric, and harmonic means over identical word pairs differ by at most 0.34 points in macro-average QA AUROC. The difference is larger on RAGTruth, where arithmetic averaging reaches 58.74 AUROC while geometric and harmonic means remain below 39. Changing the token-to-word mean has a smaller effect than changing the word-to-pair mean (Appendix \ref{app:mechanism_details}). In longer responses, more candidate pairs allow an isolated low value to dominate geometric or harmonic aggregation. Arithmetic averaging retains the neighboring confidence.

\paragraph{Words Instead of Fixed Token Windows.}
Across the 16 QA settings, LSC obtains 79.86 AUC$_s$ and 78.67 AUC$_r$, compared with 77.53 and 77.15 for the fixed two-token version. A pair of neighboring words can contain different numbers of model tokens without changing the smoothing scale. On RAGTruth, adjacent-word smoothing also improves over one-word scores, consistent with reducing the influence of an isolated unusual word. Appendix \ref{app:design_ablations} compares widths of one to five words against sentence-level confidence averaging.

\paragraph{Boundary and Continuity Tests.}
Table \ref{tab:mechanism_evidence} tests the two design choices without changing the model responses. True word boundaries outperform randomized group sizes, shifted token boundaries, and shuffled token order by 0.90, 2.50, and 3.96 AUROC points on average. With the same total reduction in log confidence, two adjacent low-confidence words increase LSC risk by 1.648 on QA and 0.629 on RAGTruth. The increase is more than three times that of either an isolated word or two separated words at the same strength. These results support boundary-aligned smoothing of contiguous confidence valleys rather than an arbitrary fixed token window. Paired bootstrap results and weaker perturbations are reported in Appendix \ref{app:mechanism_details}.

\paragraph{Paired Statistical Evidence.}
On the same QA responses, 1,000 paired bootstrap samples give LSC average gains of 17.80/14.82 AUROC points over PPL and 4.68/3.89 over Min-P under the two label protocols. After Holm correction, 26 and 19 of the 32 paired improvements over PPL and Min-P remain significant, respectively. On the 411 hallucinated RAGTruth responses, LSC also obtains the highest Token AP at $17.28\pm0.72$\%. Appendices \ref{app:mechanism_details} and \ref{app:ragtruth} provide the complete results.

\begin{figure}[tbp]
    \centering
    \includegraphics[width=\columnwidth]{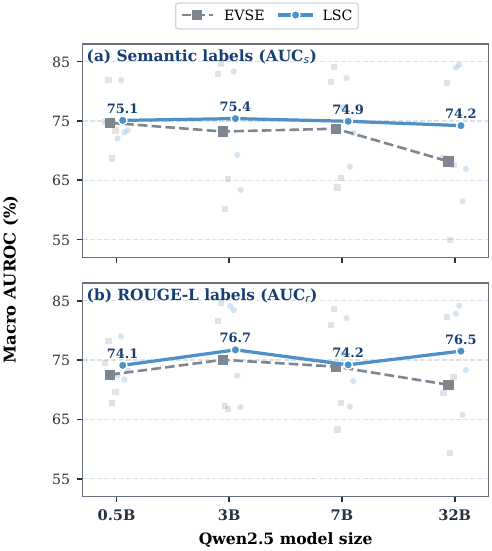}
    \caption{Performance across Qwen2.5 model sizes. Solid lines report LSC and dashed lines report EVSE. Each line is macro AUROC over NQ, TriviaQA, SQuAD, and CoQA, while faint markers show the individual datasets. LSC uses one response, whereas EVSE uses six additional responses and an auxiliary semantic model.}
    \label{fig:scaling_model_size}
\end{figure}

\paragraph{Impact of Model Size.}
Figure \ref{fig:scaling_model_size} compares Qwen2.5-Instruct models with 0.5B, 3B, 7B, and 32B parameters. The macro AUROC of LSC ranges from 74.20 to 75.38 under semantic labels and from 74.12 to 76.72 under ROUGE-L labels. Average performance is stable across model sizes, although individual datasets are not monotonic. LSC is higher than EVSE in 22 of the 32 model--dataset--label settings without requiring additional responses or an auxiliary semantic model.

\subsection{Robustness to Decoding}
The main results use greedy responses. We also vary the decoding temperature and random seed. For each of three model families, we sample 512 TriviaQA responses at temperatures $\{0.2,0.7,1.0\}$ and seeds $\{42,43,44\}$. Across all 27 settings, LSC obtains $78.04\pm6.74$ AUROC, compared with $74.43\pm7.62$ for Min-P and $76.30\pm7.40$ for the best fixed-token mean. All methods score the same responses for each temperature and seed. Appendix \ref{app:decoding_sensitivity} gives mean and variation at each temperature.

\input{sections/arxiv/cross_dataset_and_cases}

\subsection{Efficiency and Scope}
One-response probability scores share generation time. Appendix \ref{app:runtime} separates generation from sampling and NLI scoring. For 100 prompts, generating one response takes 11.2 seconds, compared with 30.7 seconds for seven responses and 55.8 seconds including NLI model loading and scoring.

LSC uses model confidence rather than external evidence, so it complements rather than replaces fact verification. Appendix \ref{app:operating_regimes} quantifies high-confidence errors, response-length effects, and variation on short conversational answers.

\input{sections/arxiv/discussion}

\section{Conclusion}
We introduced \textbf{Lowest Span Confidence (LSC)}, a zero-shot hallucination detector using one response and its output token probabilities. By selecting the lowest-confidence adjacent complete-word span, LSC captures localized uncertainty while reducing global dilution and isolated token noise. Across four model families and seven benchmarks, LSC is competitive with methods using multiple responses or model-internal information. The span analyses clarify the role of local smoothing in single-response detection without extra model calls.

\section*{Limitations}
LSC is designed as an inference-time warning signal from one response and its output token probabilities, rather than as an external fact verifier. It is most informative when a factual error produces a localized decrease in model confidence. Errors expressed with uniformly high confidence may require complementary retrieval or verification, as is also true for other confidence-based detectors. By grouping returned token pieces into complete words, LSC reduces sensitivity to isolated token splits while retaining the model's probability signal. Its localized score could support selective verification or correction of high-risk responses.

\bibliography{arxiv_custom,arxiv_extra,arxiv_multilingual}

\appendix

\captionsetup[table]{skip=3pt}
\setcounter{topnumber}{6}
\setcounter{bottomnumber}{4}
\setcounter{totalnumber}{10}
\setcounter{dbltopnumber}{4}
\renewcommand{\topfraction}{0.97}
\renewcommand{\bottomfraction}{0.90}
\renewcommand{\textfraction}{0.03}
\renewcommand{\floatpagefraction}{0.85}
\renewcommand{\dbltopfraction}{0.97}
\renewcommand{\dblfloatpagefraction}{0.85}
\setlength{\floatsep}{7pt plus 2pt minus 2pt}
\setlength{\textfloatsep}{8pt plus 2pt minus 2pt}
\setlength{\dblfloatsep}{7pt plus 2pt minus 2pt}
\setlength{\dbltextfloatsep}{8pt plus 2pt minus 2pt}

\section{Evaluation Protocols}
\label{app:evaluation_protocols}

\paragraph{Evaluation examples.}
The four QA datasets and HotpotQA use 2,000 examples per model. All detectors score the same main response. AUC$_r$ uses ROUGE-L F1 $\geq 0.5$ on NQ, TriviaQA, SQuAD, and CoQA. HotpotQA uses token F1 $\geq 0.5$.

\paragraph{Semantic correctness labels.}
For AUC$_s$, we encode each generated answer and reference with the same NLI-RoBERTa sentence encoder. We use attention-mask-aware mean pooling, lowercase the text, and limit the input to 128 tokens. A response is correct when its highest cosine similarity to a reference is at least 0.9. Similarity labels are available for all 2,000 examples in every model--dataset setting. These labels are used only for evaluation and are never given to LSC or any baseline.

\paragraph{Length normalization.}
For ordinary sequence likelihood scores, we divide negative log-likelihood by the number of valid output tokens. G-NLL is this average negative log-likelihood, and PPL is its exponential. The exponential does not change the ranking of responses, so G-NLL and PPL have identical AUROC. We normalize the likelihoods of sampled answers in the same way before using them in entropy-based baselines. We do not report unnormalized NLL for the whole sequence.

\paragraph{Statistical reporting.}
The main tables give AUROC for each set of responses. We use 1,000 paired bootstrap resamples to compare methods on the same examples and apply Holm correction across the 32 reference-based QA comparisons. The decoding analysis evaluates every temperature--seed setting for temperatures $\{0.2,0.7,1.0\}$ and seeds $\{42,43,44\}$. We report the mean $\pm$ standard deviation across seeds at each temperature.

\paragraph{Additional datasets.}
We additionally evaluate WebQuestions, PopQA, SimpleQA, and PubMedQA. Table \ref{tab:dataset_screening} reports the results and the main characteristic of each evaluation. WebQuestions and PopQA are similar to the short-answer tasks already represented by NQ and TriviaQA. SimpleQA does not have complete labels for all three model families. PubMedQA is a small classification task whose result is sensitive to answer format. We include all four results for completeness.
\input{tables/arxiv/dataset_screening_ledger}
\input{sections/arxiv/word_reconstruction}

\section{Additional Baseline Details}
\label{app:more_baseline}
Table \ref{tab:access_taxonomy} groups baselines by response budget and model access. Probability-based methods use either the probability of each generated token, $p(y_t)$, or the probabilities of all possible next tokens, $p(\cdot)$. Other methods also require hidden states or attention. Where possible, multi-response methods use the same six sampled answers. SRE and AGSER require their own generation procedures.

\paragraph{Methods Using One Response.} These methods score only the main response.
\begin{itemize}[leftmargin=*]
    \item \textbf{Perplexity (PPL)} \cite{ren2022out}: PPL is the exponential of mean negative log-likelihood over generated tokens. It can dilute a short low-confidence span within high-confidence text.

    \item \textbf{Generation NLL (G-NLL)} \cite{malinin2020uncertainty}: G-NLL is the mean negative log-likelihood over valid generated tokens. PPL is the exponential of the same value, so the two methods have identical AUROC. We use one row in the main table to avoid repeating results.

    \item \textbf{Minimum Token Probability (Min-P)} \cite{manakul2023selfcheckgpt}: Min-P uses the response's lowest selected-token probability. It can be affected by one rare word or unusual tokenization.

    \item \textbf{Energy} \cite{liu2020energy}: Energy applies log-sum-exp to next-token logits and averages the scores over the response. Unlike PPL and Min-P, it requires the full next-token distribution.

    \item \textbf{Mean Token Entropy} \cite{malinin2020uncertainty}: This baseline computes the entropy of the full next-token distribution at each decoding step and averages it over the generated response.

    \item \textbf{Meaning-Aware Response Scoring (MARS)} \cite{bakman2024mars}: MARS replaces uniform length normalization with weights from a trained importance model. It uses one response's selected-token probabilities and an auxiliary model of about 109M parameters.
\end{itemize}

\paragraph{Methods Using Several Responses.} They compare answer text and may use probabilities.
\begin{itemize}[leftmargin=*]
    \item \textbf{Lexical Similarity} \cite{kuhn2023semantic}: We average pairwise ROUGE-L F1 scores among sampled answers. Lower lexical overlap indicates less stable generations and higher uncertainty.

    \item \textbf{Semantic Entropy} \cite{farquhar2024semantic}: Sampled answers are grouped by meaning using a natural language inference (NLI) model. Their likelihood mass is aggregated within sets before computing entropy, allowing surface forms with the same meaning to be treated together.

    \item \textbf{Kernel Language Entropy (KLE)} \cite{nikitin2024kernel}: KLE represents semantic relations among sampled answers with a graph kernel and combines this structure with likelihood-weighted semantic clusters. Kernel entropy measures semantic uncertainty beyond pairwise agreement.

    \item \textbf{Semantic Graph Density (SGD)} \cite{li2025semanticgraph}: SGD measures pairwise entailment graph density among sampled answers. We weight its continuous variant by answer likelihoods.

    \item \textbf{Evidential Semantic Entropy (EVSE)} \cite{kunitomo2026evidential}: EVSE assigns evidence using sampled-answer likelihoods and semantic relations. Its entropy retains uncertainty mass beyond observed answers.

    \item \textbf{Semantic Reformulation Entropy (SRE)} \cite{tong2025sre}: We follow SRE's hybrid semantic clustering procedure with six answers, the same number used by the other multi-response baselines. The target model first produces candidate reformulations of the question. We keep the first three whose question-level SBERT cosine lies in $[0.60,0.95]$ and reject near-duplicates above $0.95$. We sample two answers for each accepted reformulation at temperature $0.8$. We then cluster the six answers in a fixed order using exact match, SBERT cosine ($\geq0.92$), and bidirectional NLI. For NLI, mean entailment must be at least $0.8$ and contradiction must be below $0.5$ in both directions. The final score is the natural-log Shannon entropy of the cluster frequencies. We do not assign a replacement score when fewer than three reformulations are valid or an additional answer is empty. Depending on the model and dataset, SRE scores 83.65\% to 99.60\% of the examples.

    \item \textbf{Length-normalized Entropy (LN-Entropy)} \cite{malinin2020uncertainty}: LN-Entropy averages per-generation predictive uncertainty after normalizing sampled answers by length. This reduces dependence on response length.
\end{itemize}

\paragraph{Methods Using Several Responses and Model Internals.} These methods also require model information not normally exposed by APIs.
\begin{itemize}[leftmargin=*]
    \item \textbf{EigenScore} \cite{chen2024inside}: EigenScore measures response dispersion through hidden-state covariance eigenvalues. Greater dispersion indicates less stable meaning across answers.

    \item \textbf{Attention-Guided Self-Reflection (AGSER)} \cite{liu2025attention}: Following the published setting, we average token contributions across layers. The two-thirds of input tokens with the largest contributions form the attentive query, and the remaining tokens form the non-attentive query. We generate one answer from each query with greedy decoding. AGSER compares both answers with the main answer using ROUGE-L ($\lambda=1$). We reverse the sign of its factuality score so that larger values indicate higher hallucination risk. AGSER requires two additional generations. In each setting, it uses the same 2,000 main responses as the other methods.

    \item \textbf{Semantically Diverse Language Generation (SDLG)} \cite{aichberger2025sdlg}: SDLG uses target-model gradients and an NLI model to steer generation toward semantically diverse alternatives. Its published evaluation uses ten generations per uncertainty estimate. We include it in the access taxonomy as a high-access method, not a one-response or API-logprob baseline.
\end{itemize}

\input{sections/arxiv/mars_comparison}

\section{RAGTruth Response and Span Evaluation}
\label{app:ragtruth}
RAGTruth provides human labels for responses and hallucinated spans in QA, summarization, and data-to-text tasks. We score its 899 test responses, of which 411 are labeled as hallucinated. We use the matching Llama-2-7B and Llama-2-13B model families to compute the probability of every token in each response, without truncating the prompt or response. LSC rebuilds complete words from the returned token pieces. Consecutive byte-fallback pieces are first decoded as UTF-8 text. We map the reconstructed words to annotated character positions. We restrict this evaluation to methods that score the released response without additional generation. Table \ref{tab:ragtruth_localization} measures the overlap between detected and human-marked hallucinated spans.
\input{tables/ragtruth_task_breakdown}
\input{tables/ragtruth_localization}

LSC has the highest Token AP overall, while the fixed-token version has slightly higher Hit@1, Hit@3, and Token IoU. We therefore use the detected span as supporting localization evidence rather than claiming exact factual-error boundaries.

\paragraph{Qualitative cases.}
RAGTruth response 12693 contains the human-annotated unsupported statement ``France opposed Indian involvement in the Vietnam War.'' LSC identifies ``opposed Indian'' as its highest-risk span. With the same model, this response lies at the 99.9th risk percentile under LSC but only the 39.5th percentile under PPL. This example shows how a local confidence valley can remain visible when whole-response averaging is less sensitive. Response 12903 illustrates a high-confidence error. Human annotators mark an unsupported paragraph giving instructions for baking hard-boiled eggs, but both LSC and PPL place the response near the first risk percentile, and the LSC span does not overlap the annotated error. Such errors require external factual verification.

\input{sections/arxiv/error_types}

\section{Mechanism and Statistical Details}
\label{app:mechanism_details}
\label{app:design_ablations}
\input{tables/arxiv/design_analysis}

Under the source-calibrated transfer protocol in Section \ref{sec:cross_dataset_transfer}, LSC obtains 69.31\% balanced accuracy, compared with 69.55\% for Min-Word, 65.69\% for fixed-token LSC, and 64.32\% for Min-P. The 0.24-point difference between the two word-level variants is small, while both transfer more reliably than the token-level alternatives.

Table \ref{tab:mechanism_full} reports three perturbation strengths. Adjacent perturbations produce the largest risk increase in five of the six dataset--strength settings. Under the mildest RAGTruth perturbation, all absolute changes in risk are below 0.021.
\input{sections/arxiv/design_ablations}

\section{Operating Regimes}
\label{app:operating_regimes}
Table \ref{tab:operating_regimes} reports length effects and high-confidence errors. LSC has lower score--length correlation than Min-Word, but its false-positive rate rises for the longest responses. All confidence-only methods have low recall on high-confidence errors.
\input{tables/operating_regimes}

\section{FELM Evaluation}
\label{app:felm}
\input{tables/felm_detailed}
FELM provides 847 open-ended responses with segment-level factuality annotations. A response is labeled as hallucinated when at least one segment is factually incorrect. Llama-2-7B, Qwen2.5-7B, and Gemma-2-9B score the same responses through teacher forcing, so the scoring model need not be the original generator. The five domains differ in sample size, so Table \ref{tab:felm_results} gives each domain equal weight. Table \ref{tab:felm_detailed} also reports pooled metrics. All three models use the same LSC definition.

\section{Detailed Results}
\label{app:detailed_results}
Tables \ref{tab:main_results} and \ref{tab:ragtruth_response} give AUROC by model. Appendix~\ref{app:ragtruth} gives task-level AUROC and AUPRC. Appendix~\ref{app:felm} gives domain-macro and pooled FELM results.

\section{Full Decoding-Sensitivity Results}
\label{app:decoding_sensitivity}
The decoding analysis varies temperature and seed, with each method using the same information. Min-P and all LSC versions use the same responses and output probabilities. Table \ref{tab:decoding_sensitivity_full} averages results over three random seeds at each temperature.
\input{tables/arxiv/decoding_sensitivity_appendix}

\section{Runtime Breakdown}
\label{app:runtime}
\input{tables/runtime_breakdown}
\paragraph{Scoring overhead and access.}
For $T$ tokens, LSC uses $O(T)$ time and memory. It needs probabilities and decoded token pieces from one response. Raw vocabulary symbols require tokenizer-specific decoding before word grouping. This adds no model calls, training, or auxiliary model.

Table \ref{tab:runtime_breakdown} separates generation from NLI scoring. NLI time without model loading is labeled ``loaded,'' and time including loading is labeled ``from start.'' Single-response methods share the same 11.2-second generation time. Six additional responses increase this to 30.7 seconds, or 55.8 seconds including NLI loading and scoring.

\section{Ethical Considerations}
\label{app:responsible_research}

\paragraph{Intended use and risks.}
LSC is a research tool for identifying responses that warrant further verification. It is not a stand-alone factuality guarantee, particularly in high-stakes settings, because models can express incorrect statements with high confidence. Deployment should combine confidence-based warnings with external evidence, task-specific calibration, and human oversight.

\paragraph{Artifacts and data.}
We use and cite public datasets and model checkpoints under their original licenses and access terms. We do not collect new human annotations or use restricted or private records. Released code and derived evaluation files will not redistribute original datasets or model weights. Users must follow these terms.

\paragraph{Use of AI assistants.}
AI assistants were used to support code development, literature search, and language polishing. All scientific claims, experimental results, analyses, citations, and final paper content were reviewed and verified by the authors.

\input{sections/arxiv/multilingual}

\end{document}

%% file: tables/arxiv/access_taxonomy.tex
\begin{table}[t]
\centering
\fontsize{5.8}{6.8}\selectfont
\setlength{\tabcolsep}{0.85pt}
\renewcommand{\arraystretch}{1.10}
\arrayrulecolor{LSCSpine}
\begin{tabular}{@{}ll|l@{}}
\toprule
\begin{tabular}[t]{@{}l@{}}\textbf{Model}\\\textbf{information}\end{tabular} &
\begin{tabular}[t]{@{}l@{}}\textbf{One response}\end{tabular} &
\begin{tabular}[t]{@{}l@{}}\textbf{Multiple responses}\end{tabular} \\
\specialrule{\lightrulewidth}{1pt}{0pt}
\begin{tabular}[t]{@{}l@{}}\textit{Output}\\\textit{text}\end{tabular} &
\multicolumn{1}{c}{\raisebox{-0.40cm}{\diagbox[width=2.78cm,height=0.58cm]{}{}}} &
\cellcolor{AccessOutput}\begin{tabular}[t]{@{}l@{}}
$+$ LexSim, ICLR'23 \namecite{kuhn2023semantic} \\
$+$ SRE$^{\dagger}$, arXiv'25 \namecite{tong2025sre}
\end{tabular} \\
\specialrule{\lightrulewidth}{1pt}{0pt}
\begin{tabular}[t]{@{}l@{}}\textit{Selected-token}\\\textit{probabilities}\\\textit{$p(y_t)$}\end{tabular} &
\cellcolor{LSCMethod}\begin{tabular}[t]{@{}l@{}}
$+$ PPL \namecite{ren2022out} \\
$+$ G-NLL \namecite{ren2022out} \\
$+$ Min-P \\
\hspace*{0.8em}\namecite{manakul2023selfcheckgpt} \\
$+$ MARS$^{\dagger}$, ACL'24 \namecite{bakman2024mars} \\
\textcolor{LSCAccent}{\textbf{$+$ LSC (ours)}}
\end{tabular} &
\cellcolor{AccessProb}\begin{tabular}[t]{@{}l@{}}
$+$ LN-Entropy, ICLR'21 \\
\hspace*{0.8em}\namecite{malinin2020uncertainty} \\
$+$ Semantic Entropy$^{\dagger}$, Nature'24 \\
\hspace*{0.8em}\namecite{farquhar2024semantic} \\
$+$ KLE$^{\dagger}$, NeurIPS'24 \namecite{nikitin2024kernel} \\
$+$ SGD$^{\dagger}$, UAI'25 \namecite{li2025semanticgraph} \\
$+$ EVSE$^{\dagger}$, EACL'26 \\
\hspace*{0.8em}\namecite{kunitomo2026evidential}
\end{tabular} \\
\specialrule{\lightrulewidth}{1pt}{0pt}
\begin{tabular}[t]{@{}l@{}}\textit{Full}\\\textit{next-token}\\\textit{distribution}\\\textit{$p(\cdot)$}\end{tabular} &
\cellcolor{AccessProb}\begin{tabular}[t]{@{}l@{}}
$+$ Energy, NeurIPS'20 \namecite{liu2020energy} \\
$+$ MTE, ICLR'21 \\
\hspace*{0.8em}\namecite{malinin2020uncertainty}
\end{tabular} &
\multicolumn{1}{c}{\raisebox{-0.40cm}{\diagbox[width=3.39cm,height=0.58cm]{}{}}} \\
\specialrule{\lightrulewidth}{1pt}{0pt}
\begin{tabular}[t]{@{}l@{}}\textit{Internal}\\\textit{states}\end{tabular} &
\multicolumn{1}{c}{\raisebox{-0.40cm}{\diagbox[width=2.78cm,height=0.58cm]{}{}}} &
\cellcolor{AccessWhite}\begin{tabular}[t]{@{}l@{}}
$+$ EigenScore, ICLR'24 \namecite{chen2024inside} \\
$+$ AGSER, EMNLP'25 \namecite{liu2025attention} \\
$+$ SDLG$^{\dagger}$, ICLR'25 \namecite{aichberger2025sdlg}
\end{tabular} \\
\bottomrule
\end{tabular}
\caption{Hallucination Detector Access Requirements. Methods are grouped by required responses and model information. LSC needs one response and selected-token probabilities. $\dagger$ means an auxiliary semantic model. Citations for PPL, G-NLL, and Min-P identify studies using these basic scores. Appendix \ref{app:more_baseline} gives method details.}
\label{tab:access_taxonomy}
\end{table}

%% file: tables/main_results.tex
\begin{table*}[!t]
\caption{Reference-Based QA Correctness. Panels (a)--(d) report AUC$_s$/AUC$_r$ (\%) for four model families on NQ, TriviaQA, SQuAD, and CoQA. \scorelegend}
\label{tab:main_results}
\centering
\fontsize{6.5}{7.3}\selectfont
\setlength{\tabcolsep}{0.8pt}
\renewcommand{\arraystretch}{0.98}
\arrayrulecolor{LSCSpine}
\begin{minipage}[t]{0.493\textwidth}
\vspace{0pt}\centering
{\fontsize{7.5}{8.3}\selectfont\bfseries (a) Llama-2-7B\par}
\vspace{1.5pt}
\begin{tabularx}{\linewidth}{@{}>{\raggedright\arraybackslash}p{0.29\linewidth}*{4}{>{\centering\arraybackslash}X}@{}}
\toprule
\textbf{Method} & \textbf{NQ} & \textbf{TriviaQA} & \textbf{SQuAD} & \textbf{CoQA} \\
\midrule
\rowcolor{AccessProb}\multicolumn{5}{l}{\textit{One response + selected-token probabilities $p(y_t)$}} \\
PPL (norm. G-NLL) & 73.56 / 73.61 & 82.45 / 81.62 & 38.79 / 47.56 & 32.28 / 37.44 \\
Min-P & \secondscore{76.13} / 75.39 & 82.01 / 81.08 & 75.16 / 77.37 & 73.12 / 76.06 \\
\lscmethodrow \textbf{LSC} & 74.63 / 73.92 & \secondscore{82.63} / 81.76 & \bestscore{83.99} / \bestscore{85.01} & \bestscore{83.47} / \bestscore{84.75} \\
\rowcolor{AccessProb}\multicolumn{5}{l}{\textit{One response + full next-token distribution $p(\cdot)$}} \\
Energy & 63.77 / 65.38 & 75.74 / 75.90 & 22.65 / 29.23 & 25.53 / 29.60 \\
Mean Token Entropy & 72.39 / 72.91 & \bestscore{83.23} / \bestscore{82.67} & 34.82 / 44.26 & 31.44 / 36.23 \\
\rowcolor{AccessOutput}\multicolumn{5}{l}{\textit{Multiple responses}} \\
Lexical Similarity & 75.14 / \secondscore{76.43} & 79.98 / 80.54 & 66.07 / 69.69 & 69.30 / 71.96 \\
LN-Entropy & 73.90 / 74.10 & 80.81 / 80.02 & 49.43 / 55.85 & 43.19 / 46.61 \\
Semantic Entropy & 75.25 / 74.71 & 79.73 / 80.17 & 75.48 / 75.17 & 80.09 / 80.39 \\
KLE & 75.60 / 75.12 & 80.24 / 80.45 & 75.51 / 75.15 & 78.72 / 79.69 \\
SGD & 76.04 / 75.40 & 80.47 / 80.27 & 76.08 / 75.78 & 78.79 / 79.97 \\
EVSE & \bestscore{76.96} / \bestscore{76.46} & 82.50 / \secondscore{82.20} & \secondscore{78.36} / \secondscore{78.47} & \secondscore{81.62} / \secondscore{81.53} \\
SRE & 70.36 / 69.39 & 74.86 / 74.33 & 59.78 / 59.60 & 62.68 / 61.73 \\
\rowcolor{AccessInternal}\multicolumn{5}{l}{\textit{Multiple responses + internal states}} \\
EigenScore & 73.89 / 72.99 & 78.50 / 77.60 & 73.69 / 75.97 & 70.88 / 71.79 \\
AGSER & 67.09 / 66.34 & 69.89 / 71.34 & 22.70 / 27.04 & 57.36 / 57.26 \\
\bottomrule
\end{tabularx}
\end{minipage}
\hfill
\begin{minipage}[t]{0.493\textwidth}
\vspace{0pt}\centering
{\fontsize{7.5}{8.3}\selectfont\bfseries (b) Qwen2.5-7B\par}
\vspace{1.5pt}
\centering
\begin{tabularx}{\linewidth}{@{}>{\raggedright\arraybackslash}p{0.29\linewidth}*{4}{>{\centering\arraybackslash}X}@{}}
\toprule
\textbf{Method} & \textbf{NQ} & \textbf{TriviaQA} & \textbf{SQuAD} & \textbf{CoQA} \\
\midrule
\rowcolor{AccessProb}\multicolumn{5}{l}{\textit{One response + selected-token probabilities $p(y_t)$}} \\
PPL (norm. G-NLL) & 75.47 / 74.53 & 83.17 / 82.58 & 33.62 / 46.77 & 30.39 / 41.23 \\
Min-P & 79.71 / 78.98 & 83.63 / 83.14 & 54.50 / 59.79 & 57.02 / 60.17 \\
\lscmethodrow \textbf{LSC} & 77.27 / 76.11 & 82.21 / 82.06 & \bestscore{67.28} / \bestscore{67.15} & \bestscore{72.96} / \bestscore{71.48} \\
\rowcolor{AccessProb}\multicolumn{5}{l}{\textit{One response + full next-token distribution $p(\cdot)$}} \\
Energy & 58.65 / 57.70 & 76.21 / 75.84 & 31.28 / 40.21 & 26.25 / 31.14 \\
Mean Token Entropy & 74.20 / 73.35 & \bestscore{84.11} / \bestscore{83.81} & 32.93 / 46.15 & 29.18 / 40.30 \\
\rowcolor{AccessOutput}\multicolumn{5}{l}{\textit{Multiple responses}} \\
Lexical Similarity & 78.82 / 79.49 & 81.03 / 81.34 & 49.32 / 58.58 & 49.03 / 56.33 \\
LN-Entropy & 76.29 / 75.36 & 81.76 / 81.19 & 44.51 / 53.62 & 42.05 / 50.28 \\
Semantic Entropy & 79.39 / 78.98 & 80.65 / 80.98 & 60.13 / 61.36 & 60.08 / 62.99 \\
KLE & 79.96 / 79.84 & 81.15 / 81.42 & 62.08 / 63.03 & 61.77 / 64.32 \\
SGD & \secondscore{80.52} / \secondscore{80.35} & 81.86 / 81.54 & \secondscore{64.23} / \secondscore{64.32} & 61.77 / 64.33 \\
EVSE & \bestscore{81.50} / \bestscore{80.92} & \secondscore{84.04} / \secondscore{83.55} & 63.79 / 63.29 & \secondscore{65.40} / \secondscore{67.81} \\
SRE & 70.49 / 70.16 & 79.14 / 79.03 & 51.93 / 55.59 & 47.61 / 50.16 \\
\rowcolor{AccessInternal}\multicolumn{5}{l}{\textit{Multiple responses + internal states}} \\
EigenScore & 79.69 / 78.46 & 80.41 / 79.60 & 57.16 / 60.76 & 56.32 / 61.56 \\
AGSER & 58.96 / 60.18 & 72.24 / 75.40 & 49.90 / 57.54 & 51.32 / 57.15 \\
\bottomrule
\end{tabularx}
\end{minipage}
\par\vspace{5pt}\noindent
\begin{minipage}[t]{0.493\textwidth}
\vspace{0pt}\centering
{\fontsize{7.5}{8.3}\selectfont\bfseries (c) Gemma-2-9B\par}
\vspace{1.5pt}
\centering
\begin{tabularx}{\linewidth}{@{}>{\raggedright\arraybackslash}p{0.29\linewidth}*{4}{>{\centering\arraybackslash}X}@{}}
\toprule
\textbf{Method} & \textbf{NQ} & \textbf{TriviaQA} & \textbf{SQuAD} & \textbf{CoQA} \\
\midrule
\rowcolor{AccessProb}\multicolumn{5}{l}{\textit{One response + selected-token probabilities $p(y_t)$}} \\
PPL (norm. G-NLL) & 55.63 / 56.09 & 79.89 / 80.40 & 62.67 / 65.72 & 58.10 / 60.14 \\
Min-P & 69.13 / 70.75 & 78.49 / 79.60 & \secondscore{77.32} / \secondscore{77.81} & 67.38 / 67.98 \\
\lscmethodrow \textbf{LSC} & \bestscore{79.26} / \bestscore{78.88} & \bestscore{83.56} / \secondscore{83.18} & \bestscore{80.00} / \bestscore{79.12} & \bestscore{73.98} / \bestscore{73.69} \\
\rowcolor{AccessProb}\multicolumn{5}{l}{\textit{One response + full next-token distribution $p(\cdot)$}} \\
Energy & 63.61 / 63.09 & 55.32 / 54.38 & 73.16 / 70.48 & 61.48 / 58.41 \\
Mean Token Entropy & 61.71 / 61.25 & 80.10 / 80.65 & 56.02 / 59.48 & 56.36 / 58.61 \\
\rowcolor{AccessOutput}\multicolumn{5}{l}{\textit{Multiple responses}} \\
Lexical Similarity & 70.06 / \secondscore{72.14} & \secondscore{80.92} / \bestscore{83.93} & 69.18 / 73.70 & 66.21 / 69.20 \\
LN-Entropy & 64.63 / 65.99 & 79.99 / 81.39 & 59.23 / 63.31 & 56.13 / 59.71 \\
Semantic Entropy & 67.22 / 68.63 & 76.65 / 77.65 & 72.77 / 75.05 & 70.62 / 71.47 \\
KLE & 69.04 / 70.28 & 78.62 / 80.17 & 73.72 / 75.42 & 70.62 / 71.72 \\
SGD & \secondscore{70.57} / 71.22 & 80.07 / 81.39 & 74.52 / 76.05 & \secondscore{70.68} / \secondscore{71.93} \\
EVSE & 69.46 / 70.68 & 78.02 / 80.22 & 74.17 / 76.60 & 69.74 / 71.46 \\
SRE & 66.49 / 66.57 & 75.38 / 75.85 & 68.42 / 68.04 & 63.42 / 64.59 \\
\rowcolor{AccessInternal}\multicolumn{5}{l}{\textit{Multiple responses + internal states}} \\
EigenScore & 63.77 / 65.36 & 77.10 / 78.06 & 74.58 / 76.55 & 67.86 / 69.74 \\
AGSER & 64.51 / 63.75 & 54.70 / 56.13 & 50.75 / 53.82 & 65.20 / 67.71 \\
\bottomrule
\end{tabularx}
\end{minipage}
\hfill
\begin{minipage}[t]{0.493\textwidth}
\vspace{0pt}\centering
{\fontsize{7.5}{8.3}\selectfont\bfseries (d) Mistral-7B-v0.3\par}
\vspace{1.5pt}
\centering
\begin{tabularx}{\linewidth}{@{}>{\raggedright\arraybackslash}p{0.29\linewidth}*{4}{>{\centering\arraybackslash}X}@{}}
\toprule
\textbf{Method} & \textbf{NQ} & \textbf{TriviaQA} & \textbf{SQuAD} & \textbf{CoQA} \\
\midrule
\rowcolor{AccessProb}\multicolumn{5}{l}{\textit{One response + selected-token probabilities $p(y_t)$}} \\
PPL (norm. G-NLL) & 76.07 / 69.21 & 79.66 / 76.58 & 64.16 / 63.57 & 66.99 / 64.64 \\
Min-P & 87.11 / 77.78 & \secondscore{88.79} / \secondscore{85.37} & 75.30 / 72.11 & \secondscore{77.99} / 73.10 \\
\lscmethodrow \textbf{LSC} & \bestscore{89.40} / \bestscore{83.53} & \bestscore{89.44} / \bestscore{87.73} & \secondscore{80.42} / \secondscore{76.48} & 77.22 / 73.91 \\
\rowcolor{AccessProb}\multicolumn{5}{l}{\textit{One response + full next-token distribution $p(\cdot)$}} \\
Energy & 68.44 / 64.67 & 68.62 / 66.97 & 64.36 / 63.72 & 66.30 / 65.37 \\
Mean Token Entropy & 77.93 / 70.48 & 78.85 / 76.09 & 63.14 / 63.04 & 66.63 / 64.89 \\
\rowcolor{AccessOutput}\multicolumn{5}{l}{\textit{Multiple responses}} \\
Lexical Similarity & 70.04 / 68.77 & 78.44 / 77.87 & 66.09 / 67.73 & 73.26 / 72.53 \\
LN-Entropy & 77.17 / 72.39 & 77.91 / 75.94 & 64.53 / 64.89 & 65.69 / 64.49 \\
Semantic Entropy & 84.90 / 80.04 & 78.67 / 78.54 & 64.59 / 63.95 & 70.47 / 69.91 \\
KLE & 86.60 / 81.58 & 81.31 / 80.55 & 66.73 / 65.48 & 72.47 / 71.30 \\
SGD & \secondscore{87.61} / 82.78 & 83.36 / 82.08 & 68.76 / 67.72 & 74.61 / 72.89 \\
EVSE & 87.18 / \secondscore{82.88} & 85.09 / 83.89 & 70.22 / 69.39 & 76.11 / \secondscore{74.87} \\
SRE & 77.87 / 69.59 & 72.56 / 71.80 & 61.09 / 58.43 & 67.67 / 66.92 \\
\rowcolor{AccessInternal}\multicolumn{5}{l}{\textit{Multiple responses + internal states}} \\
EigenScore & 74.80 / 73.21 & 71.42 / 71.73 & 61.63 / 63.35 & 59.63 / 60.67 \\
AGSER & 24.67 / 33.44 & 46.90 / 47.51 & \bestscore{82.12} / \bestscore{79.14} & \bestscore{78.26} / \bestscore{75.59} \\
\bottomrule
\end{tabularx}
\end{minipage}
\end{table*}

%% file: tables/broader_results.tex
\begin{table}[!t]
\caption{Human-Annotated Factuality on RAGTruth. AUROC (\%) is reported for each scorer and for all 899 responses. Released responses cover QA, summarization, and data-to-text generation. Methods requiring new samples or internal states are not reported. \scorelegend}
\label{tab:ragtruth_response}
\centering
\fontsize{7.5}{8.5}\selectfont
\setlength{\tabcolsep}{3.1pt}
\renewcommand{\arraystretch}{1.10}
\arrayrulecolor{LSCSpine}
\begin{tabularx}{\columnwidth}{l*{3}{>{\centering\arraybackslash}X}}
\toprule
\textbf{Method} & \textbf{Llama-7B} & \textbf{Llama-13B} & \textbf{All} \\
\midrule
\rowcolor{AccessProb}\multicolumn{4}{l}{\textit{One response + selected-token probabilities $p(y_t)$}} \\
PPL (normalized G-NLL) & \secondscore{50.73} & \secondscore{43.53} & \secondscore{47.32} \\
Min-P & 46.30 & 33.70 & 42.82 \\
\lscmethodrow \textbf{LSC} & \bestscore{56.12} & \bestscore{61.16} & \bestscore{58.74} \\
\bottomrule
\end{tabularx}
\end{table}

%% file: tables/felm_results.tex
\begin{table}[t]
\caption{Open-Ended Factuality on FELM. AUROC (\%) is macro-averaged over five domains. Each model scores the same 847 released responses by teacher forcing. \scorelegend}
\label{tab:felm_results}
\centering
\fontsize{7.5}{8.5}\selectfont
\setlength{\tabcolsep}{3.4pt}
\renewcommand{\arraystretch}{1.08}
\arrayrulecolor{LSCSpine}
\begin{tabularx}{\columnwidth}{l*{4}{>{\centering\arraybackslash}X}}
\toprule
\textbf{Method} & \textbf{Llama} & \textbf{Qwen} & \textbf{Gemma} & \textbf{Avg.} \\
\midrule
\rowcolor{AccessProb}\multicolumn{5}{l}{\textit{Selected-token probabilities $p(y_t)$}} \\
PPL (normalized G-NLL) & 60.76 & 57.85 & 53.19 & 57.27 \\
Min-P & 53.85 & 46.25 & 46.92 & 49.00 \\
\lscmethodrow \textbf{LSC} & \secondscore{60.84} & 60.02 & 57.77 & 59.55 \\
\rowcolor{AccessProb}\multicolumn{5}{l}{\textit{Full next-token distribution $p(\cdot)$}} \\
Energy & \bestscore{61.39} & \bestscore{65.42} & \bestscore{67.79} & \bestscore{64.87} \\
Mean Token Entropy & 60.75 & \secondscore{61.92} & \secondscore{66.79} & \secondscore{63.15} \\
\bottomrule
\end{tabularx}
\end{table}

%% file: tables/hotpot_results.tex
\begin{table}[t]
\caption{Multi-Hop QA Correctness on HotpotQA. AUROC (\%) is reported for greedy responses labeled incorrect when token F1 is below 0.5. \scorelegend}
\label{tab:hotpot_results}
\centering
\fontsize{7.5}{8.5}\selectfont
\setlength{\tabcolsep}{2.5pt}
\renewcommand{\arraystretch}{1.04}
\arrayrulecolor{LSCSpine}
\begin{tabular}{lccc}
\toprule
\textbf{Method} & \textbf{Llama-2-7B} & \textbf{Qwen2.5-7B} & \textbf{Gemma-2-9B} \\
\midrule
\rowcolor{AccessProb}\multicolumn{4}{l}{\textit{One response + selected-token probabilities $p(y_t)$}} \\
PPL (normalized G-NLL) & 56.96 & 73.79 & 65.14 \\
Min-P & \secondscore{63.18} & \secondscore{77.26} & 61.34 \\
\lscmethodrow \textbf{LSC} & \bestscore{65.66} & \bestscore{79.92} & \bestscore{69.80} \\
\rowcolor{AccessProb}\multicolumn{4}{l}{\textit{One response + full next-token distribution $p(\cdot)$}} \\
Energy & 54.01 & 66.00 & 45.56 \\
Mean Token Entropy & 56.85 & 74.51 & \secondscore{67.66} \\
\bottomrule
\end{tabular}
\end{table}

%% file: tables/arxiv/mechanism_evidence.tex
\begin{table*}[t]
\caption{Mechanism Tests. Panel (a) averages 16 QA settings after changing only the grouping or order of identical probability traces. Panel (b) reports LSC risk increases under the same total log-confidence reduction at $\alpha=e^{-2}$. The largest and second-largest values are shown in red/bold and blue/underlined.}
\label{tab:mechanism_evidence}
\centering
\fontsize{7.5}{8.5}\selectfont
\begin{minipage}[t]{\columnwidth}
\centering
\textbf{(a) Boundary and order placebo}\par\vspace{2pt}
\setlength{\tabcolsep}{7pt}
\begin{tabularx}{\linewidth}{l*{2}{>{\centering\arraybackslash}X}}
\toprule
\textbf{Grouping} & \textbf{AUC$_s$} & \textbf{AUC$_r$} \\
\midrule
Randomized group sizes & \secondscore{78.83} & \secondscore{77.90} \\
Shifted token boundaries & 76.96 & 76.57 \\
Shuffled token order & 75.68 & 74.93 \\
\lscmethodrow \textbf{True word boundaries} & \bestscore{79.86} & \bestscore{78.67} \\
\bottomrule
\end{tabularx}
\end{minipage}
\hfill
\begin{minipage}[t]{\columnwidth}
\centering
\textbf{(b) Controlled confidence perturbation}\par\vspace{2pt}
\setlength{\tabcolsep}{6pt}
\begin{tabularx}{\linewidth}{l*{2}{>{\centering\arraybackslash}X}}
\toprule
\textbf{Perturbed words} & \textbf{QA} & \textbf{RAGTruth} \\
\midrule
One isolated & \secondscore{0.494} & \secondscore{0.162} \\
Two separated & 0.423 & 0.134 \\
\lscmethodrow \textbf{Two adjacent} & \bestscore{1.648} & \bestscore{0.629} \\
\bottomrule
\end{tabularx}
\end{minipage}
\end{table*}

%% file: sections/arxiv/cross_dataset_and_cases.tex
\subsection{Detection under Dataset Shift}
\label{sec:cross_dataset_transfer}

AUROC evaluates ranking but does not specify a threshold for binary detection. We select a threshold on one source dataset by maximizing Youden's $J$ statistic and apply it unchanged to the other three QA datasets. Repeating this for four model families and both label protocols gives 96 directed transfers per method. Target labels are used only for evaluation. Balanced accuracy weights classes equally.

\begin{figure*}[t]
    \centering
    \includegraphics[width=\textwidth]{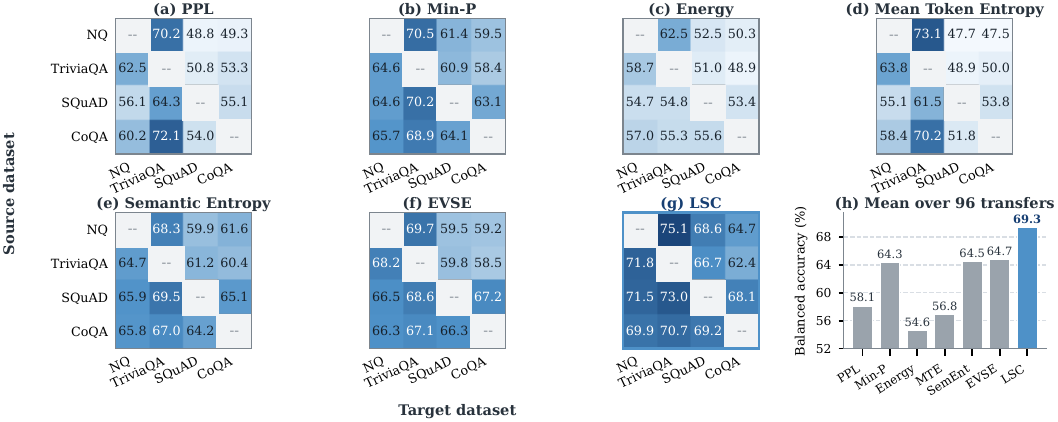}
    \caption{Cross-dataset threshold transfer. Rows denote the source dataset used to select the threshold, and columns denote the target dataset. Panels (a)--(g) report PPL, Min-P, Energy, Mean Token Entropy, Semantic Entropy, EVSE, and LSC. Each heatmap cell averages balanced accuracy (\%) over four model families and two label protocols. The diagonal is omitted. Panel (h) averages all 96 directed transfers per method. Methods score between 99.2\% and 100\% of responses.}
    \label{fig:cross_dataset_transfer}
\end{figure*}

Among the methods in Figure \ref{fig:cross_dataset_transfer}, LSC retains 69.31\% balanced accuracy after transfer. This exceeds EVSE at 64.75\%, Semantic Entropy at 64.47\%, Min-P at 64.32\%, and the remaining probability-based scores. The improvement holds without target-dataset calibration and is largest when transferring to SQuAD and CoQA. LSC requires only one response and its output token probabilities, whereas Semantic Entropy and EVSE use multiple responses. Appendix \ref{app:mechanism_details} compares word- and token-level scores under the same protocol.

The same pattern holds for MCC. LSC reaches 0.376, compared with 0.308 for EVSE, 0.299 for Semantic Entropy, and 0.293 for Min-P. LSC also has the smallest standard deviation in balanced accuracy across the 96 transfers at 0.062. Its mean balanced accuracy is 70.11\% under semantic labels and 68.51\% under ROUGE-L labels. Source--target averages range from 62.4\% to 75.1\%, so transfer still depends on the source and target datasets.

\subsection{Case Study and Failure Cases}
\label{sec:case_study}
\input{tables/qualitative_cases}

Table \ref{tab:qualitative_cases} illustrates the scope of the confidence signal. In the first case, an unsupported relation creates a local confidence valley. The selected pair overlaps the human annotation even though the whole response remains relatively confident. In the second case, a confident but unsupported cooking procedure gives LSC no useful signal. Localized uncertainty can guide further inspection, while uniformly confident errors require external evidence.

%% file: tables/qualitative_cases.tex
\begin{table*}[t]
\caption{Representative RAGTruth Cases. Bold text marks the word pair selected by LSC. Risk percentiles are computed within the corresponding scorer and task, with larger values indicating higher hallucination risk.}
\label{tab:qualitative_cases}
\centering
\small
\setlength{\tabcolsep}{5pt}
\renewcommand{\arraystretch}{1.12}
\begin{tabularx}{\textwidth}{@{}>{\raggedright\arraybackslash}p{0.16\textwidth}XX@{}}
\toprule
& \textbf{Localized error detected} & \textbf{Overconfident error missed} \\
\midrule
\textbf{Question} & Difference between Vietnam and India & How to bake hard-boiled eggs \\
\textbf{Generated claim} & ``France \textbf{opposed Indian} involvement in the Vietnam War.'' & ``Bake the eggs in a preheated oven at 300\,$^{\circ}$F for 15--20 minutes.'' \\
\textbf{Human label} & Baseless information. The passages instead state that India opposed American involvement. & Baseless information. The passages describe boiling eggs and give no oven instructions. \\
\textbf{PPL risk} & 0.37 \,(39.5th percentile) & 0.15 \,(1.0th percentile) \\
\textbf{LSC risk} & \textbf{5.89} \,(\textbf{99.9th percentile}) & \textbf{0.69} \,(\textbf{1.0th percentile}) \\
\textbf{Selected span} & ``\textbf{opposed Indian}'' \quad \ding{51} overlaps the human span & ``\textbf{Based on}'' \quad \ding{55} does not overlap the human span \\
\bottomrule
\end{tabularx}
\end{table*}

%% file: sections/arxiv/discussion.tex
\subsection{Discussion}
\label{sec:discussion}

The results identify two practical regimes. When an error creates a contiguous confidence valley, LSC raises response-level risk and identifies a suspicious span, as in Table \ref{tab:qualitative_cases}. Averaging reduces the influence of an isolated unusual word. A uniformly confident error may leave no local probability signal. Retrieval or external verification can complement confidence-based detection in these cases.

Word boundaries provide an intermediate unit between model tokens and the full response. A pair can contain a variable number of tokens without assuming that a factual claim spans two words. The width and sentence-level comparisons in Appendix \ref{app:design_ablations} separate this local scoring choice from semantic claim boundaries. Appendix \ref{app:multilingual} examines the same whitespace rule on multilingual responses, where an unspaced stretch remains one surface unit, without assuming a language-specific word boundary.

Cross-dataset transfer connects the design to deployment. LSC remains useful when a threshold selected on one dataset is applied to another, including against detectors using complete token distributions or multiple responses. It can therefore support a fixed decision rule before target labels are available, while allowing later recalibration as the task or response distribution changes.

%% file: tables/arxiv/dataset_screening_ledger.tex
\begin{table*}[!t]
\caption{Additional Dataset Evaluation. Values are average AUROC (\%) over three model families when complete labels are available. SimpleQA reports the one model family with complete attempted-answer labels. \scorelegend}
\label{tab:dataset_screening}
\centering
\fontsize{7.5}{8.5}\selectfont
\setlength{\tabcolsep}{4pt}
\renewcommand{\arraystretch}{1.06}
\begin{tabular*}{\textwidth}{@{\extracolsep{\fill}}llllrr@{}}
\toprule
\textbf{Dataset} & \textbf{Examples} & \textbf{Evaluation label} & \textbf{Dataset characteristic} & \textbf{LSC} & \textbf{Best baseline} \\
\midrule
WebQuestions & 2,000 & token F1 $<0.5$ & Short-answer QA & \bestscore{72.52} & \secondscore{71.08} (Min-P) \\
PopQA & 2,000 & token F1 $<0.5$ & Short-answer QA & \secondscore{81.42} & \bestscore{81.97} (Min-P) \\
SimpleQA & 1,972 & Kimi-K2.6 correctness labels & One scoring model & \secondscore{51.80} & \bestscore{61.51} (Min-P) \\
PubMedQA & 500 & incorrect yes/no/maybe answer & Three-way classification & \secondscore{64.83} & \bestscore{66.10} (Energy) \\
\bottomrule
\end{tabular*}
\end{table*}

%% file: sections/arxiv/word_reconstruction.tex
\subsection{Word Reconstruction}
\label{app:word_reconstruction}

\paragraph{Surface units and token overlap.}
We define a surface word as a maximal non-whitespace interval $[a_k,b_k)$ in the decoded response. Whitespace includes spaces, line breaks, tabs, and Unicode whitespace. Suppose token $i$ occupies character interval $[s_i,e_i)$. It overlaps word $k$ when $e_i>a_k$ and $s_i<b_k$. Let $I_k$ contain these overlapping token indices. The group in Eq.~(1) is
\begin{equation}
G_k=\{\min I_k,\ldots,\max I_k\}.
\end{equation}
Thus, a token that crosses a word boundary can contribute to both words. The groups need not be a disjoint partition of model tokens. Whitespace-only pieces between the first and last overlapping tokens stay in that range, while leading or trailing whitespace alone does not create a word.

\paragraph{Decoding before grouping.}
For SentencePiece and byte-level BPE, word grouping uses decoded text rather than raw vocabulary symbols. Leading-space markers indicate whitespace, not part of a word. Consecutive byte-fallback pieces are decoded together as UTF-8, retaining the mapping from each decoded character to its contributing tokens. Raw byte-level BPE symbols require the tokenizer's byte decoder, since replacing space markers alone does not recover all non-ASCII text. APIs that return decoded pieces or token bytes provide the same information without another model forward pass. Algorithm \ref{alg:arxiv_reconstruction} summarizes grouping.

\begin{algorithm}[t]
\caption{Reconstructing surface-word groups.}
\label{alg:arxiv_reconstruction}
\small
\begin{algorithmic}[1]
\Require Token pieces or bytes $z_{1:T}$ and their decoding rule
\Ensure Decoded text $y$ and word groups $G_{1:M}$
\State Decode $z_{1:T}$, preserving token--character offsets
\State Find maximal non-whitespace intervals $[a_k,b_k)$ in $y$
\For{each interval $[a_k,b_k)$ in text order}
    \State Find tokens with $e_i>a_k$ and $s_i<b_k$
    \State $G_k\gets\{\text{first through last overlapping token}\}$
\EndFor
\State \Return $y,G_{1:M}$
\end{algorithmic}
\end{algorithm}

\paragraph{Punctuation and clitics.}
Attached punctuation stays in the same surface word: ``Paris,'' forms one group, while ``Paris ,'' forms two. Apostrophes and hyphens do not create boundaries without whitespace. Arabic clitics likewise remain attached, since the method does not perform morphological segmentation. A stretch of Chinese, Japanese, or Thai text without whitespace forms one surface unit. Appendix \ref{app:multilingual} evaluates this convention separately from the English QA study. The score applies to responses with at least one surface unit, using $-\log q_1$ for a single-unit response.

\paragraph{Detection under different token representations.}
Table \ref{tab:arxiv_reconstruction} compares the original reconstruction with full tokenizer decoding on 31,940 QA responses with available token probabilities. The answers, probabilities, and correctness labels remain unchanged. Replacing encoded whitespace markers with their literal characters preserves the scores. Full decoding changes the AUC$_s$ and AUC$_r$ averages by less than 0.01 points. The table reports each tokenization condition separately, with overlapping categories. These sparse categories do not support a general claim about all writing systems.

\input{tables/arxiv/word_reconstruction}

One whitespace-only QA response retains its token-based score in Table \ref{tab:arxiv_reconstruction}. Excluding it leaves overall AUROC unchanged to two decimal places.

%% file: tables/arxiv/word_reconstruction.tex
\begin{table*}[t]
\captionsetup{justification=raggedright,singlelinecheck=false}
\caption{Reconstruction comparison on the original QA responses. Values are AUC$_s$/AUC$_r$, averaged equally over model--dataset settings containing both label classes. Conditions overlap.}
\label{tab:arxiv_reconstruction}
\centering
\fontsize{8.5}{10}\selectfont
\setlength{\tabcolsep}{4pt}
\renewcommand{\arraystretch}{1.10}
\arrayrulecolor{LSCSpine}
\begin{tabular*}{\textwidth}{@{\extracolsep{\fill}}lrrrr@{}}
\toprule
\textbf{Condition} & \textbf{Responses} & \textbf{LSC} & \textbf{Decoded-text LSC} & \textbf{Fixed two-token} \\
\midrule
All QA responses & 31,940 & 79.86 / 78.67 & 79.86 / 78.67 & 77.53 / 77.15 \\
Multi-token word & 26,470 & 80.47 / 78.95 & 80.47 / 78.96 & 77.52 / 77.06 \\
Attached punctuation & 19,422 & 77.24 / 76.84 & 77.24 / 76.84 & 74.82 / 75.43 \\
Separate punctuation & 265 & 91.94 / 78.07 & 91.94 / 78.07 & 92.26 / 79.45 \\
Apostrophe or hyphen & 4,563 & 83.07 / 79.23 & 83.07 / 79.23 & 80.71 / 78.08 \\
SentencePiece-style tokens & 23,940 & 81.50 / 80.16 & 81.50 / 80.16 & 79.29 / 78.34 \\
Byte-level BPE tokens & 8,000 & 74.93 / 74.20 & 74.94 / 74.20 & 72.23 / 73.60 \\
Whitespace-only pieces & 14,795 & 81.93 / 78.59 & 81.93 / 78.60 & 78.28 / 75.84 \\
Line break & 15,788 & 79.65 / 79.26 & 79.66 / 79.26 & 75.39 / 75.13 \\
Consecutive whitespace & 1,770 & 81.15 / 75.05 & 81.15 / 75.05 & 76.46 / 66.35 \\
Byte-fallback tokens & 5,119 & 79.27 / 82.88 & 79.27 / 82.88 & 70.41 / 69.47 \\
Non-ASCII response text & 619 & 89.39 / 84.03 & 89.39 / 84.03 & 86.83 / 84.68 \\
\bottomrule
\end{tabular*}
\par\vspace{3pt}
\begin{minipage}{\textwidth}
Counts include all responses in each category. Under semantic/ROUGE-L labels, separate punctuation contributes 110/228 responses, line breaks 15,754/15,782, consecutive whitespace 1,756/1,756, byte fallback 5,082/5,114, and non-ASCII text 580/580. Byte fallback includes ASCII characters such as line breaks.
\end{minipage}
\end{table*}

%% file: sections/arxiv/mars_comparison.tex
\subsection{Comparison with MARS}
\label{app:mars_comparison}
We compare LSC with the single-response Confidence + MARS variant of \citet{bakman2024mars}, using the authors' released \texttt{duygunuryldz/MARS} BERT importance predictor with approximately 109M parameters. Following the original scoring rule, the risk score combines global mean token negative log-likelihood and importance-weighted mean phrase negative log-likelihood with equal weights of 0.5. Phrase importance uses normalized sigmoid outputs from the predictor. We use the released weights without additional training or parameter selection on the evaluation data.

The importance predictor receives the question and response. For SQuAD and CoQA, the passage and dialogue history remain part of the target LLM's generation prompt and hence its conditional token probabilities, but are not added to the importance predictor's input. These results evaluate the question--response importance model on context-dependent answers and open-domain QA. Both methods score the same responses with the original token probabilities and correctness labels. Table~\ref{tab:arxiv_mars_comparison} gives the response count for each comparison. We exclude 60 Gemma responses without usable probability sequences, one empty Llama response and one Qwen response whose importance weights could not be aligned to tokens.

\input{tables/arxiv/mars_comparison}

Across the four models and four datasets, LSC achieves mean $\mathrm{AUC}_s/\mathrm{AUC}_r$ of 79.86/78.67, compared with 64.09/65.95 for Confidence + MARS. On the open-domain NQ and TriviaQA datasets alone, the corresponding scores are 82.30/80.90 and 76.58/75.62. MARS is slightly stronger on TriviaQA with Llama-2 and Qwen2.5, while LSC has higher scores on the remaining comparisons. These results show that local confidence aggregation can provide competitive single-response detection without an auxiliary importance model.

%% file: tables/arxiv/mars_comparison.tex
\begin{table*}[t]
\captionsetup{justification=raggedright,singlelinecheck=false}
\caption{Comparison with Confidence + MARS on four QA benchmarks. Entries give $\mathrm{AUC}_s/\mathrm{AUC}_r$ (\%) on the same $N$ responses. The mean weights each model--dataset setting equally. Bold marks the higher scores.}
\label{tab:arxiv_mars_comparison}
\centering
\fontsize{8.5}{10}\selectfont
\setlength{\tabcolsep}{4pt}
\renewcommand{\arraystretch}{1.10}
\arrayrulecolor{LSCSpine}
\begin{tabular*}{\textwidth}{@{\extracolsep{\fill}}llrrr@{}}
\toprule
\textbf{Model} & \textbf{Dataset} & $\boldsymbol{N}$ & \textbf{LSC} & \textbf{Confidence + MARS} \\
\midrule
\multirow{4}{*}{Llama-2-7B} & NQ & 2,000 & \textbf{74.63 / 73.92} & 73.53 / 73.75 \\
& TriviaQA & 2,000 & 82.63 / 81.76 & \textbf{82.68 / 82.07} \\
& SQuAD & 2,000 & \textbf{83.99 / 85.01} & 47.96 / 55.19 \\
& CoQA & 1,999 & \textbf{83.46 / 84.74} & 41.18 / 44.63 \\
\midrule
\multirow{4}{*}{Qwen2.5-7B} & NQ & 2,000 & \textbf{77.27 / 76.11} & 75.12 / 73.86 \\
& TriviaQA & 2,000 & 82.21 / 82.06 & \textbf{83.41 / 82.85} \\
& SQuAD & 2,000 & \textbf{67.28 / 67.15} & 35.96 / 48.38 \\
& CoQA & 1,999 & \textbf{72.95 / 71.46} & 34.10 / 44.11 \\
\midrule
\multirow{4}{*}{Gemma-2-9B} & NQ & 1,986 & \textbf{79.26 / 78.88} & 62.33 / 63.38 \\
& TriviaQA & 1,954 & \textbf{83.56 / 83.18} & 80.20 / 80.97 \\
& SQuAD & 2,000 & \textbf{80.00 / 79.12} & 62.34 / 66.19 \\
& CoQA & 2,000 & \textbf{73.98 / 73.69} & 60.23 / 62.46 \\
\midrule
\multirow{4}{*}{Mistral-7B} & NQ & 2,000 & \textbf{89.40 / 83.53} & 76.20 / 70.95 \\
& TriviaQA & 2,000 & \textbf{89.44 / 87.73} & 79.17 / 77.13 \\
& SQuAD & 2,000 & \textbf{80.42 / 76.48} & 63.99 / 64.08 \\
& CoQA & 2,000 & \textbf{77.22 / 73.91} & 67.10 / 65.19 \\
\midrule
\multicolumn{2}{l}{\textbf{Mean}} & & \textbf{79.86 / 78.67} & 64.09 / 65.95 \\
\bottomrule
\end{tabular*}
\end{table*}

%% file: tables/ragtruth_task_breakdown.tex
\begin{table*}[!t]
\caption{RAGTruth Response-Level Results by Task. Each entry is AUROC/AUPRC (\%). ``All'' pools the 899 responses and is not an average of the three task columns. \scorelegend}
\label{tab:ragtruth_tasks}
\centering
\fontsize{7.4}{8.3}\selectfont
\setlength{\tabcolsep}{6.5pt}
\begin{tabularx}{\textwidth}{l*{4}{>{\centering\arraybackslash}X}}
\toprule
\textbf{Method} & \textbf{QA} & \textbf{Summarization} & \textbf{Data-to-text} & \textbf{All} \\
\midrule
PPL (normalized G-NLL) & 48.43 / 24.35 & 48.45 / 27.38 & 25.83 / 75.09 & 47.32 / 41.52 \\
Min-P & 41.03 / 23.78 & \bestscore{50.72} / \bestscore{28.44} & 38.76 / 77.77 & 42.82 / 41.16 \\
Fixed-token LSC & \secondscore{59.29} / \secondscore{32.66} & \secondscore{50.59} / 26.05 & 43.95 / 81.81 & \secondscore{56.78} / \secondscore{47.90} \\
Min-Word & 43.71 / 24.43 & 48.42 / \secondscore{28.03} & \secondscore{44.23} / \secondscore{82.08} & 37.49 / 38.83 \\
\lscmethodrow \textbf{LSC} & \bestscore{62.41} / \bestscore{35.71} & 50.26 / 27.33 & \bestscore{48.73} / \bestscore{83.26} & \bestscore{58.74} / \bestscore{50.29} \\
\bottomrule
\end{tabularx}
\end{table*}

%% file: tables/ragtruth_localization.tex
\begin{table*}[!t]
\caption{Hallucination-Span Localization on RAGTruth. Values are mean $\pm$ standard deviation (\%) over 1,000 bootstrap samples of 411 hallucinated responses. Hit@$k$ tests span overlap, Token IoU measures overlap, and Token AP ranks annotated error tokens. \scorelegend}
\label{tab:ragtruth_localization}
\centering
\fontsize{7.4}{8.4}\selectfont
\setlength{\tabcolsep}{2.4pt}
\begin{tabularx}{\textwidth}{>{\raggedright\arraybackslash}X*{4}{>{\centering\arraybackslash}X}}
\toprule
\textbf{Method} & \textbf{Hit@1} & \textbf{Hit@3} & \textbf{Token IoU} & \textbf{Token AP} \\
\midrule
Min-P & 0.98$\pm$0.46 & 24.52$\pm$2.12 & 0.02$\pm$0.01 & 13.97$\pm$0.56 \\
Fixed-token LSC & \bestscore{14.84$\pm$1.75} & \bestscore{37.67$\pm$2.41} & \bestscore{1.40$\pm$0.23} & \secondscore{16.18$\pm$0.64} \\
Min-Word & 4.72$\pm$1.05 & 26.07$\pm$2.17 & 0.23$\pm$0.06 & 15.40$\pm$0.59 \\
\lscmethodrow LSC & \secondscore{14.30$\pm$1.68} & \secondscore{35.54$\pm$2.36} & \secondscore{1.37$\pm$0.21} & \bestscore{17.28$\pm$0.72} \\
\bottomrule
\end{tabularx}
\end{table*}

%% file: sections/arxiv/error_types.tex
\paragraph{Detection by error type.}
Table \ref{tab:ragtruth_error_types} separates RAGTruth responses by the annotation categories. For each category, responses containing at least one error of that type are compared with the same 488 hallucination-free responses. Responses with only other error types are excluded. A response may contain several types, so the positive groups overlap. Relative to PPL, LSC improves AUROC by 13.29 points for Evident Baseless Info and 12.79 points for Evident Conflict. The two subtle-error groups contain only 27 and eight positive responses, respectively, and provide less precise estimates. These labels describe the errors, not the model's confidence, so this analysis is separate from the high-confidence evaluation in Appendix \ref{app:operating_regimes}.

\input{tables/arxiv/error_types}

%% file: tables/arxiv/error_types.tex
\begin{table}[t]
\captionsetup{justification=raggedright,singlelinecheck=false}
\caption{RAGTruth AUROC (\%) by error type. Each positive group is compared with the same 488 hallucination-free responses. Groups overlap. Bold marks the highest AUROC.}
\label{tab:ragtruth_error_types}
\centering
\fontsize{8.5}{10}\selectfont
\setlength{\tabcolsep}{2.5pt}
\renewcommand{\arraystretch}{1.10}
\arrayrulecolor{LSCSpine}
\begin{tabular*}{\columnwidth}{@{\extracolsep{\fill}}lrrrr@{}}
\toprule
\textbf{Error type} & $N_+$ & \textbf{PPL} & \textbf{Min-P} & \textbf{LSC} \\
\midrule
Evident Baseless Info & 254 & 45.38 & 41.81 & \textbf{58.67} \\
Evident Conflict & 215 & 47.32 & 43.37 & \textbf{60.11} \\
Subtle Baseless Info & 27 & 41.86 & 37.27 & \textbf{58.98} \\
Subtle Conflict & 8 & 51.08 & \textbf{63.86} & 56.56 \\
\bottomrule
\end{tabular*}
\end{table}

%% file: tables/arxiv/design_analysis.tex
\begin{table*}[!t]
\begin{minipage}[t]{\textwidth}
\captionof{table}{Aggregation and Span Analysis. QA values average 16 settings across four model families and four datasets. The three word-pair variants use identical word groups and differ only in their mean. RAGTruth reports AUROC on all 899 human-annotated responses. All methods use one response and selected-token probabilities. \scorelegend}
\label{tab:design_analysis}
\centering
\fontsize{7.3}{8.2}\selectfont
\setlength{\tabcolsep}{5.0pt}
\renewcommand{\arraystretch}{1.06}
\begin{tabularx}{\textwidth}{>{\raggedright\arraybackslash}X>{\raggedright\arraybackslash}X>{\raggedright\arraybackslash}Xrrr}
\toprule
\textbf{Method} & \textbf{Aggregation} & \textbf{Candidate span} & \textbf{QA AUC$_s$} & \textbf{QA AUC$_r$} & \textbf{RAGTruth} \\
\midrule
Min-P & minimum & one token & 75.17 & 74.78 & 42.82 \\
Fixed-token LSC & arithmetic mean & two tokens & 77.53 & 77.15 & \secondscore{56.78} \\
Min-Word & minimum & one word & \bestscore{80.70} & 78.61 & 37.49 \\
Geometric LSC & geometric mean & two words & 80.07 & \bestscore{78.75} & 38.33 \\
Harmonic LSC & harmonic mean & two words & \secondscore{80.19} & \secondscore{78.73} & 37.55 \\
\lscmethodrow \textbf{LSC} & arithmetic mean & two words & 79.86 & 78.67 & \bestscore{58.74} \\
\bottomrule
\end{tabularx}
\end{minipage}

\vspace{7pt}
\input{tables/arxiv/mechanism_full}
\hfill
\input{tables/arxiv/statistical_robustness}
\end{table*}

%% file: tables/arxiv/mechanism_full.tex
\begin{minipage}[t]{\columnwidth}
\vspace{0pt}
\captionof{table}{Confidence-Perturbation Stress Test. Values are mean increases in LSC risk. For each $\alpha$, the total log-confidence reduction is the same across the three perturbations. \scorelegend}
\label{tab:mechanism_full}
\centering
\fontsize{7.2}{8.1}\selectfont
\setlength{\tabcolsep}{4.5pt}
\begin{tabularx}{\columnwidth}{>{\centering\arraybackslash}p{0.12\columnwidth}>{\raggedright\arraybackslash}X*{2}{>{\centering\arraybackslash}X}}
\toprule
\textbf{$\alpha$} & \textbf{Perturbed words} & \textbf{QA} & \textbf{RAGTruth} \\
\midrule
\multirow{3}{*}{$e^{-0.5}$} & One isolated & \secondscore{0.199} & \bestscore{0.021} \\
 & Two separated & 0.115 & 0.016 \\
 & Two adjacent & \bestscore{0.243} & \secondscore{0.019} \\
\midrule
\multirow{3}{*}{$e^{-1}$} & One isolated & \secondscore{0.374} & \secondscore{0.073} \\
 & Two separated & 0.241 & 0.047 \\
 & Two adjacent & \bestscore{0.666} & \bestscore{0.100} \\
\midrule
\multirow{3}{*}{$e^{-2}$} & One isolated & \secondscore{0.494} & \secondscore{0.162} \\
 & Two separated & 0.423 & 0.134 \\
 & Two adjacent & \bestscore{1.648} & \bestscore{0.629} \\
\bottomrule
\end{tabularx}
\end{minipage}

%% file: tables/arxiv/statistical_robustness.tex
\begin{minipage}[t]{\columnwidth}
\vspace{0pt}
\captionof{table}{Paired Statistical Comparison on Reference-Based QA. $\Delta$ is LSC minus the baseline in AUROC points. The final columns count positive gains and Holm-adjusted $p<0.05$ among 32 comparisons.}
\label{tab:paired_statistics}
\centering
\fontsize{7.8}{9}\selectfont
\setlength{\tabcolsep}{3pt}
\renewcommand{\arraystretch}{1.08}
\begin{tabular*}{\linewidth}{@{\extracolsep{\fill}}lrrrr@{}}
\toprule
\textbf{Baseline} & \textbf{$\Delta_s$} & \textbf{$\Delta_r$} & \textbf{$\Delta>0$} & \textbf{Holm} \\
\midrule
PPL & +17.80 & +14.82 & 30 & 26 \\
Min-P & +4.68 & +3.89 & 25 & 19 \\
Fixed-token LSC & +2.33 & +1.52 & 18 & 13 \\
Min-Word & $-$0.84 & +0.07 & 12 & 3 \\
\bottomrule
\end{tabular*}
\end{minipage}

%% file: sections/arxiv/design_ablations.tex
\subsection{Aggregation at Both Levels}
\label{app:two_level_aggregation}

We vary token-to-word and word-to-pair aggregation separately while keeping the responses, token probabilities, word groups, and labels unchanged. In Table~\ref{tab:arxiv_two_level_means}, the first letter denotes the token-to-word mean and the second denotes the word-to-pair mean. A and G indicate arithmetic and geometric averaging. All variants select the lowest pair confidence and apply the same negative-log transformation. We use 31,940 QA responses with available probabilities, 899 RAGTruth responses, and 847 FELM responses. QA results average the four model families and four datasets equally. FELM results average the five domains within each scoring model, then average the three models.

\input{tables/arxiv/design_means}

Changing only the word-to-pair mean from arithmetic to geometric reduces RAGTruth AUROC by 20.40 points with arithmetic token-to-word aggregation and by 19.47 points with geometric token-to-word aggregation. Changing only the token-to-word mean has a much smaller effect. Thus, most of this difference comes from smoothing adjacent word confidence rather than averaging subword pieces. When one word confidence approaches zero, the arithmetic pair mean retains half of the neighboring confidence, whereas the geometric mean approaches zero. The QA differences are small, while FELM favors GA over AA. Arithmetic averaging is a simple smoothing choice, not necessarily the best mean for every dataset.

\subsection{Word-Span Width and Adjacency}
\label{app:word_width_adjacency}

We compare contiguous windows of one to five words using the same arithmetic word confidence. For a response containing $M<n$ words, the window contains all $M$ words. The two-word window exceeds the best wider window, $n=3$, by 1.74 AUC$_s$ points and 1.16 AUC$_r$ points on QA (Table~\ref{tab:arxiv_word_width}). The one-word score is higher under AUC$_s$, while wider windows improve RAGTruth and FELM. These results distinguish a fixed local smoothing scale from the length of a factual claim.

\input{tables/arxiv/design_width}

To isolate adjacency, we also average the two lowest-confidence words anywhere in the response. The two positions must be distinct but may be adjacent. For $M=1$, both methods use the only word. This alternative changes only the permitted pair locations. LSC improves RAGTruth AUROC by 15.06 points and FELM domain-macro AUROC by 1.13 points. The unrestricted pair is stronger on QA, indicating that the value of adjacency depends on the response structure rather than simply on including two low-confidence words.

We further compare hallucinated and hallucination-free RAGTruth responses from the same task and scoring model, restricting their word-count ratio to $[0.8,1.25]$. Over the resulting 8,142 response pairs, LSC attains 52.31\% pairwise concordance, compared with 43.43\% for the unrestricted pair. Concordance counts the fraction of pairs in which the hallucinated response receives higher risk, with half credit for ties. The 8.88-point difference remains when task, scoring model, and approximate response length are controlled, but is not the pooled AUROC reported above.

\subsection{Sentence-Level Aggregation}
\label{app:sentence_aggregation}

The sentence-level baseline uses English Punkt sentence segmentation~\citep{kiss-strunk-2006-unsupervised} from NLTK 3.9.2, averages word confidence within each sentence, and selects the lowest sentence mean. We map sentence endpoints to complete-word endpoints and merge an isolated numeric list marker, such as ``1.'', into the following sentence. The same text rules apply to all datasets. Punkt is a statistical sentence segmenter, not a large language model or a factual-claim extractor. No detector is trained.

\input{tables/arxiv/design_sentences}

Table~\ref{tab:arxiv_sentence_aggregation} uses the same responses for both methods within each dataset. The QA comparison contains 31,937 responses whose text and word groups can be aligned. It excludes the 60 responses without token probabilities and three additional responses without the required alignment. LSC improves average AUC$_s$/AUC$_r$ by 11.16/9.44 points on QA and domain-macro AUROC by 1.53 points on FELM. Sentence averaging is stronger on RAGTruth overall. Within RAGTruth, LSC versus sentence averaging is 62.41 versus 53.70 on QA, 50.26 versus 53.95 on summarization, and 48.73 versus 60.42 on data-to-text. This comparison concerns the scale of confidence aggregation, not recovery of true factual boundaries.

%% file: tables/arxiv/design_means.tex
\begin{table}[t]
\captionsetup{justification=raggedright,singlelinecheck=false}
\caption{Two-level aggregation. A/G denotes arithmetic/geometric averaging within words and then within pairs. AUROC (\%) averages 16 settings for QA and five domains and three scorers for FELM.}
\label{tab:arxiv_two_level_means}
\centering
\fontsize{8.5}{10}\selectfont
\setlength{\tabcolsep}{2.5pt}
\renewcommand{\arraystretch}{1.10}
\arrayrulecolor{LSCSpine}
\begin{tabularx}{\columnwidth}{>{\raggedright\arraybackslash}Xrrrr}
\toprule
\textbf{Mean} & \shortstack{\textbf{QA}\\AUC$_s$} & \shortstack{\textbf{QA}\\AUC$_r$} & \textbf{RAGTruth} & \textbf{FELM} \\
\midrule
\lscmethodrow AA (LSC) & 79.86 & 78.67 & 58.74 & 59.55 \\
GA & 79.67 & 78.60 & 57.21 & 60.41 \\
AG & 80.07 & 78.75 & 38.33 & 58.40 \\
GG & 79.93 & 78.73 & 37.74 & 57.43 \\
\bottomrule
\end{tabularx}
\end{table}

%% file: tables/arxiv/design_width.tex
\begin{table}[t]
\captionsetup{justification=raggedright,singlelinecheck=false}
\caption{Word-span width and adjacency. Windows use $\min(n,M)$ words. ``Any two'' averages the two lowest-confidence words anywhere, including adjacent positions. AUROC (\%) is summarized as in Table~\ref{tab:arxiv_two_level_means}.}
\label{tab:arxiv_word_width}
\centering
\fontsize{8.5}{10}\selectfont
\setlength{\tabcolsep}{2.5pt}
\renewcommand{\arraystretch}{1.10}
\arrayrulecolor{LSCSpine}
\begin{tabularx}{\columnwidth}{>{\raggedright\arraybackslash}Xrrrr}
\toprule
\textbf{Window} & \shortstack{\textbf{QA}\\AUC$_s$} & \shortstack{\textbf{QA}\\AUC$_r$} & \textbf{RAGTruth} & \textbf{FELM} \\
\midrule
$n=1$ & 80.70 & 78.61 & 37.49 & 56.54 \\
\lscmethodrow $n=2$ (LSC) & 79.86 & 78.67 & 58.74 & 59.55 \\
$n=3$ & 78.12 & 77.51 & 59.90 & 60.24 \\
$n=4$ & 76.39 & 76.18 & 60.68 & 61.15 \\
$n=5$ & 74.84 & 74.92 & 61.62 & 61.51 \\
\midrule
Any two & 82.03 & 80.42 & 43.68 & 58.42 \\
\bottomrule
\end{tabularx}
\end{table}

%% file: tables/arxiv/design_sentences.tex
\begin{table}[t]
\captionsetup{justification=raggedright,singlelinecheck=false}
\caption{Local and sentence-level aggregation. QA entries give AUC$_s$/AUC$_r$ over four model families. RAGTruth uses pooled AUROC and FELM domain-macro AUROC over three scorers. $N$ counts responses, not repeated FELM scores.}
\label{tab:arxiv_sentence_aggregation}
\centering
\fontsize{8.5}{10}\selectfont
\setlength{\tabcolsep}{2.5pt}
\renewcommand{\arraystretch}{1.10}
\arrayrulecolor{LSCSpine}
\begin{tabular*}{\columnwidth}{@{\extracolsep{\fill}}lrrr@{}}
\toprule
\textbf{Dataset} & $N$ & \textbf{LSC} & \textbf{Sentence} \\
\midrule
NQ & 7,986 & 80.14/78.11 & 73.94/71.88 \\
TriviaQA & 7,954 & 84.46/83.68 & 81.98/81.03 \\
SQuAD & 8,000 & 77.92/76.94 & 59.56/63.06 \\
CoQA & 7,997 & 76.85/75.92 & 59.24/60.95 \\
\midrule
RAGTruth & 899 & 58.74 & 67.06 \\
FELM & 847 & 59.55 & 58.01 \\
\bottomrule
\end{tabular*}
\end{table}

%% file: tables/operating_regimes.tex
\begin{table*}[!t]
\caption{Operating Regimes of Confidence-Only Detectors. Panel (a) reports the mean Spearman correlation between risk and response word count among correct QA responses, followed by false-positive rate at 80\% TPR in the shortest and longest length quartiles. Panel (b) reports RAGTruth recall at a threshold set to 10\% FPR. A high-confidence hallucination has PPL-NLL no greater than the median for correct responses from the same model. Rankings follow the preferred direction of each metric. \scorelegend}
\label{tab:operating_regimes}
\centering
\fontsize{7.3}{8.2}\selectfont
\begin{minipage}[t]{\columnwidth}
\centering
\textbf{(a) Response-length dependence}\par\vspace{2pt}
\setlength{\tabcolsep}{4.5pt}
\begin{tabularx}{\linewidth}{l*{3}{>{\centering\arraybackslash}X}}
\toprule
\textbf{Method} & \textbf{$\rho$} & \textbf{FPR Q1} & \textbf{FPR Q4} \\
\midrule
PPL & \bestscore{$-$0.083} & 57.18 & \bestscore{53.52} \\
Min-P & 0.219 & 39.86 & \secondscore{66.55} \\
Fixed-token LSC & \secondscore{0.195} & 38.25 & 67.24 \\
Min-Word & 0.430 & \bestscore{24.58} & 76.89 \\
\lscmethodrow LSC & 0.284 & \secondscore{29.59} & 71.49 \\
\bottomrule
\end{tabularx}
\end{minipage}
\hfill
\begin{minipage}[t]{\columnwidth}
\centering
\textbf{(b) Overconfident hallucinations}\par\vspace{2pt}
\setlength{\tabcolsep}{5.5pt}
\begin{tabularx}{\linewidth}{l*{2}{>{\centering\arraybackslash}X}}
\toprule
\textbf{Method} & \textbf{High conf.} & \textbf{Other} \\
\midrule
PPL & 0.00 & 7.66 \\
Min-P & 3.57 & 12.66 \\
Fixed-token LSC & \secondscore{5.92} & \bestscore{15.00} \\
Min-Word & 3.04 & 13.01 \\
\lscmethodrow LSC & \bestscore{9.41} & \secondscore{14.71} \\
\bottomrule
\end{tabularx}
\end{minipage}
\end{table*}

%% file: tables/felm_detailed.tex
\begin{table*}[!t]
\caption{Detailed FELM Results Averaged Across Three Scoring Models. Domain macro averages the five domains equally. Pooled metrics treat all 847 responses as one set. Values are percentages. \scorelegend}
\label{tab:felm_detailed}
\centering
\fontsize{7.5}{8.5}\selectfont
\setlength{\tabcolsep}{5pt}
\begin{tabularx}{\textwidth}{>{\raggedright\arraybackslash}X*{4}{>{\centering\arraybackslash}X}}
\toprule
\textbf{Method} & \textbf{Dom. AUC} & \textbf{Dom. AP} & \textbf{Pool AUC} & \textbf{Pool AP} \\
\midrule
PPL & 57.27 & 41.67 & 58.18 & 40.95 \\
Min-P & 49.00 & 35.15 & 44.46 & 29.97 \\
Fixed-token LSC & 54.81 & 38.31 & 52.52 & 35.12 \\
Min-Word & 56.54 & 40.95 & 51.44 & 34.45 \\
\lscmethodrow LSC & 59.55 & 43.07 & 56.48 & 39.41 \\
\midrule
Energy & \bestscore{64.87} & \bestscore{49.49} & \bestscore{62.35} & \secondscore{44.74} \\
Mean Token Entropy & \secondscore{63.15} & \secondscore{48.90} & \secondscore{61.29} & \bestscore{44.75} \\
\bottomrule
\end{tabularx}
\end{table*}

%% file: tables/arxiv/decoding_sensitivity_appendix.tex
\begin{table*}[!t]
\caption{Robustness to Sampled Decoding. AUROC (\%) is mean $\pm$ standard deviation over seeds 42, 43, and 44 for 512 sampled TriviaQA responses per row. All methods score the same responses. \scorelegend}
\label{tab:decoding_sensitivity_full}
\centering
\fontsize{7.0}{8.0}\selectfont
\setlength{\tabcolsep}{4.4pt}
\begin{tabularx}{\textwidth}{ll*{5}{>{\centering\arraybackslash}X}}
\toprule
\textbf{Model} & \textbf{Temp.} & \textbf{Min-P} & \textbf{Fixed $w=2$} & \textbf{Fixed $w=3$} & \textbf{Fixed $w=4$} & \textbf{LSC} \\
\midrule
\multirow{3}{*}{Gemma-2-9B}
& 0.2 & 69.10 $\pm$ 0.08 & 66.55 $\pm$ 0.72 & 70.27 $\pm$ 0.23 & \secondscore{74.96 $\pm$ 0.66} & \bestscore{79.37 $\pm$ 0.08} \\
& 0.7 & 69.44 $\pm$ 2.66 & 69.52 $\pm$ 1.24 & 73.36 $\pm$ 0.76 & \secondscore{76.39 $\pm$ 0.30} & \bestscore{80.81 $\pm$ 0.83} \\
& 1.0 & 72.05 $\pm$ 3.15 & 72.46 $\pm$ 0.79 & 75.77 $\pm$ 1.02 & \secondscore{78.33 $\pm$ 1.41} & \bestscore{82.03 $\pm$ 1.04} \\
\midrule
\multirow{3}{*}{Llama-2-7B}
& 0.2 & \secondscore{67.68 $\pm$ 1.74} & \bestscore{67.71 $\pm$ 1.37} & 67.47 $\pm$ 1.27 & 64.98 $\pm$ 1.18 & 66.62 $\pm$ 0.65 \\
& 0.7 & 67.34 $\pm$ 3.47 & \secondscore{68.81 $\pm$ 2.98} & \bestscore{68.96 $\pm$ 2.92} & 67.09 $\pm$ 2.78 & 68.58 $\pm$ 1.75 \\
& 1.0 & 71.33 $\pm$ 5.15 & \bestscore{72.88 $\pm$ 4.13} & \secondscore{72.60 $\pm$ 4.08} & 71.29 $\pm$ 3.79 & 72.27 $\pm$ 2.02 \\
\midrule
\multirow{3}{*}{Qwen2.5-7B}
& 0.2 & 83.24 $\pm$ 0.15 & \bestscore{83.85 $\pm$ 0.25} & \secondscore{83.72 $\pm$ 0.32} & 83.13 $\pm$ 0.32 & 82.87 $\pm$ 0.21 \\
& 0.7 & 84.09 $\pm$ 0.22 & \bestscore{84.76 $\pm$ 0.29} & \secondscore{84.63 $\pm$ 0.63} & 84.12 $\pm$ 0.59 & 83.83 $\pm$ 0.52 \\
& 1.0 & 85.61 $\pm$ 2.51 & \bestscore{86.92 $\pm$ 1.89} & \secondscore{86.81 $\pm$ 1.80} & 86.43 $\pm$ 1.73 & 85.93 $\pm$ 1.46 \\
\bottomrule
\end{tabularx}
\end{table*}

%% file: tables/runtime_breakdown.tex
\begin{table*}[!t]
\caption{Runtime Breakdown. Times are measured on 100 NQ questions with Qwen2.5-7B and one NVIDIA L20X. Multi-response methods use one main response and six additional samples.}
\label{tab:runtime_breakdown}
\centering
\fontsize{8.5}{9.5}\selectfont
\setlength{\tabcolsep}{3.0pt}
\begin{tabularx}{\textwidth}{>{\raggedright\arraybackslash}X*{2}{>{\centering\arraybackslash}p{0.16\textwidth}}}
\toprule
\textbf{Stage} & \textbf{Responses} & \textbf{Time (s)} \\
\midrule
Main response + probabilities & 1 & 11.2 \\
Six additional samples & 6 & 19.6 \\
All response generation & 7 & 30.7 \\
NLI scoring (model loaded) & -- & 8.5 \\
NLI scoring (including model loading) & -- & 25.1 \\
Total including NLI model loading & 7 & 55.8 \\
\bottomrule
\end{tabularx}
\end{table*}

%% file: sections/arxiv/multilingual.tex
\section{Multilingual Evaluation}
\label{app:multilingual}
\input{tables/arxiv/multilingual}

\paragraph{Datasets and labels.}
We evaluate human-annotated responses in six languages. Mu-SHROOM \citep{vazquez2025mushroom} provides character-level annotations. We evaluate its English, German, Chinese, and Arabic test sets, labeling a response positive if its majority-vote annotation contains a hallucinated span. JHARS \citep{kamei2025jhars} contains Japanese responses to 150 questions from three generators. We use its relaxed annotations, retaining 443 responses with resolved sentence labels. A response is positive if any sentence is contradictory or unverifiable. The stricter annotation setting leaves 423 responses and is reported separately. HAT \citep{chatterjee2026hat} provides professional translation annotations. We use all 2,699 English-to-Thai test translations with their original binary hallucination labels.

\paragraph{Scoring protocol.}
Qwen2.5-7B-Instruct and Gemma-2-9B-it score each released response through teacher forcing. Mu-SHROOM uses the released question. For JHARS, a fixed Japanese instruction asks the model to answer using only the reference passage, followed by the passage and question. HAT uses the instruction ``Translate the following English text into Thai. Return only the translation.'' followed by the English source. The chat template supplies the response prefix. We compute selected-token probabilities by normalizing over the full vocabulary, without truncating responses or using hallucination labels as model inputs. Probabilities come from the scoring model, which may differ from the response generator.

\paragraph{Surface units and comparisons.}
The scoring formula is unchanged. Whitespace separates surface units, and an unspaced stretch remains one unit rather than being divided into language-specific words. PPL, Min-P, fixed two-token averaging, and LSC use the same probabilities. Table \ref{tab:multilingual_response} reports response-level AUROC rather than character localization. LSC uses the original token-piece reconstruction. Applying the same rule to fully decoded text leaves reported AUROC unchanged except for Qwen2.5 on Chinese, from 75.06 to 74.60. We evaluate all 150 Arabic responses with both models. No language-specific detector is trained.

\paragraph{Results.}
On Thai HAT, LSC exceeds Min-P by 8.85 AUROC points and fixed two-token averaging by 4.98 points, averaged over both models. PPL has the highest AUROC in both Thai comparisons. On Japanese JHARS, LSC averages 69.06 AUROC under the relaxed annotations, while the stronger baseline depends on the scoring model. Mu-SHROOM results vary by language and model. Each language set has only three to seven hallucination-free responses, making AUROC sensitive to individual examples. These task-specific results do not establish a common ranking across languages.

\paragraph{Language-specific segmentation.}
We examine whether segmentation changes local scoring units. On Japanese JHARS, 229 of the 443 responses form one whitespace unit and 60 form two, leaving no choice among multiple adjacent pairs in those responses. Replacing whitespace grouping with SudachiPy, its core dictionary, and SplitMode.A changes LSC AUROC from 67.13 to 70.84 for Qwen2.5 and from 70.99 to 74.29 for Gemma-2. In contrast, PyThaiNLP with the default \texttt{newmm} segmenter gives 55.36 and 56.38 on HAT, below whitespace-based LSC. Both variants retain surface forms and punctuation and leave the aggregation rule unchanged. Explicit segmentation can change detection performance, but these comparisons do not support a universal segmentation improvement.

\paragraph{Data use.}
Mu-SHROOM uses the CC BY 4.0 license and HAT uses CC BY-NC-ND 4.0. The JHARS repository is licensed under Apache 2.0.

%% file: tables/arxiv/multilingual.tex
\begin{table*}[t]
\captionsetup{justification=raggedright,singlelinecheck=false}
\caption{Response-level AUROC (\%) on human-annotated multilingual data. $N_+$/$N_-$ count hallucinated/hallucination-free responses. Fixed-token averages adjacent token pairs. Qwen2.5-7B and Gemma-2-9B score released responses by teacher forcing. Bold marks the highest AUROC.}
\label{tab:multilingual_response}
\centering
\fontsize{8.5}{10}\selectfont
\setlength{\tabcolsep}{4pt}
\renewcommand{\arraystretch}{1.10}
\arrayrulecolor{LSCSpine}
\begin{tabular*}{\textwidth}{@{\extracolsep{\fill}}llrrrrrr@{}}
\toprule
\textbf{Dataset / language} & \textbf{Scorer} & $N_+$ & $N_-$ & \textbf{PPL} & \textbf{Min-P} & \textbf{Fixed-token} & \textbf{LSC} \\
\midrule
Mu-SHROOM / English & Qwen2.5 & 149 & 5 & 39.60 & 40.40 & \textbf{64.30} & 61.21 \\
 & Gemma-2 & 149 & 5 & 37.58 & \textbf{63.49} & 59.87 & 51.01 \\
\addlinespace[2pt]
Mu-SHROOM / German & Qwen2.5 & 146 & 4 & 41.78 & 45.38 & 63.70 & \textbf{65.07} \\
 & Gemma-2 & 146 & 4 & 39.55 & 31.34 & 53.25 & \textbf{53.94} \\
\addlinespace[2pt]
Mu-SHROOM / Chinese & Qwen2.5 & 147 & 3 & 65.08 & 48.53 & 59.18 & \textbf{75.06} \\
 & Gemma-2 & 147 & 3 & 67.35 & \textbf{78.68} & 55.78 & 73.24 \\
\addlinespace[2pt]
Mu-SHROOM / Arabic & Qwen2.5 & 143 & 7 & 45.35 & 35.26 & 40.26 & \textbf{48.95} \\
 & Gemma-2 & 143 & 7 & 49.55 & 28.57 & 55.34 & \textbf{59.34} \\
\midrule
JHARS / Japanese & Qwen2.5 & 28 & 415 & 69.35 & 65.82 & \textbf{72.22} & 67.13 \\
(relaxed annotations) & Gemma-2 & 28 & 415 & 72.40 & \textbf{74.38} & 73.89 & 70.99 \\
\addlinespace[2pt]
JHARS / Japanese & Qwen2.5 & 11 & 412 & 72.88 & 70.08 & \textbf{75.66} & 62.95 \\
(strict annotations) & Gemma-2 & 11 & 412 & \textbf{73.39} & 69.13 & 72.99 & 66.26 \\
\midrule
HAT / Thai & Qwen2.5 & 102 & 2,597 & \textbf{69.37} & 58.54 & 63.54 & 67.67 \\
 & Gemma-2 & 102 & 2,597 & \textbf{70.04} & 61.16 & 63.90 & 69.73 \\
\bottomrule
\end{tabular*}
\end{table*}

%% file: arxiv_custom.bib
@inproceedings{chen2024inside,
  title = {{INSIDE}: {LLM}s' Internal States Retain the Power of Hallucination Detection},
  author = {Chen, Chao and
      Liu, Kai and
      Chen, Ze and
      Gu, Yi and
      Wu, Yue and
      Tao, Mingyuan and
      Fu, Zhihang and
      Ye, Jieping},
  booktitle = {International Conference on Learning Representations},
  year = {2024},
  pages = {3056--3076},
  url = {https://proceedings.iclr.cc/paper_files/paper/2024/hash/0d1986a61e30e5fa408c81216a616e20-Abstract-Conference.html}
}

@inproceedings{zhang2023sac3,
  title = {{SAC}$^3$: Reliable Hallucination Detection in Black-Box Language Models via Semantic-aware Cross-check Consistency},
  author = {Zhang, Jiaxin and
      Li, Zhuohang and
      Das, Kamalika and
      Malin, Bradley and
      Kumar, Sricharan},
  booktitle = {Findings of the Association for Computational Linguistics: EMNLP 2023},
  year = {2023},
  pages = {15445--15458},
  publisher = {Association for Computational Linguistics},
  doi = {10.18653/v1/2023.findings-emnlp.1032},
  url = {https://aclanthology.org/2023.findings-emnlp.1032/}
}

@inproceedings{kuhn2023semantic,
  title = {Semantic Uncertainty: Linguistic Invariances for Uncertainty Estimation in Natural Language Generation},
  author = {Kuhn, Lorenz and
      Gal, Yarin and
      Farquhar, Sebastian},
  booktitle = {International Conference on Learning Representations},
  year = {2023},
  url = {https://iclr.cc/virtual/2023/poster/11013}
}

@misc{tong2025sre,
  title = {Semantic Reformulation Entropy for Robust Hallucination Detection in {QA} Tasks},
  author = {Chaodong Tong and
      Qi Zhang and
      Lei Jiang and
      Yanbing Liu and
      Nannan Sun and
      Wei Li},
  year = {2025},
  eprint = {2509.17445},
  archivePrefix = {arXiv},
  primaryClass = {cs.CL},
  url = {https://arxiv.org/abs/2509.17445}
}

@inproceedings{li2024dawn,
  title = {The Dawn After the Dark: An Empirical Study on Factuality Hallucination in Large Language Models},
  author = {Li, Junyi and
      Chen, Jie and
      Ren, Ruiyang and
      Cheng, Xiaoxue and
      Zhao, Xin and
      Nie, Jian-Yun and
      Wen, Ji-Rong},
  booktitle = {Proceedings of the 62nd Annual Meeting of the Association for Computational Linguistics (Volume 1: Long Papers)},
  year = {2024},
  pages = {10879--10899},
  publisher = {Association for Computational Linguistics},
  doi = {10.18653/v1/2024.acl-long.586},
  url = {https://aclanthology.org/2024.acl-long.586/}
}

@inproceedings{manakul2023selfcheckgpt,
  title = {{S}elf{C}heck{GPT}: Zero-Resource Black-Box Hallucination Detection for Generative Large Language Models},
  author = {Manakul, Potsawee and
      Liusie, Adian and
      Gales, Mark},
  booktitle = {Proceedings of the 2023 Conference on Empirical Methods in Natural Language Processing},
  year = {2023},
  pages = {9004--9017},
  publisher = {Association for Computational Linguistics},
  doi = {10.18653/v1/2023.emnlp-main.557},
  url = {https://aclanthology.org/2023.emnlp-main.557/}
}

@inproceedings{he2024llm,
  title = {{LLM} {Factoscope}: Uncovering {LLM}s' Factual Discernment through Measuring Inner States},
  author = {He, Jinwen and
      Gong, Yujia and
      Lin, Zijin and
      Wei, Cheng{'}an and
      Zhao, Yue and
      Chen, Kai},
  booktitle = {Findings of the Association for Computational Linguistics: ACL 2024},
  year = {2024},
  pages = {10218--10230},
  publisher = {Association for Computational Linguistics},
  doi = {10.18653/v1/2024.findings-acl.608},
  url = {https://aclanthology.org/2024.findings-acl.608/}
}

@inproceedings{chuang2024lookback,
  title = {{Lookback Lens}: Detecting and Mitigating Contextual Hallucinations in Large Language Models Using Only Attention Maps},
  author = {Chuang, Yung-Sung and
      Qiu, Linlu and
      Hsieh, Cheng-Yu and
      Krishna, Ranjay and
      Kim, Yoon and
      Glass, James R.},
  booktitle = {Proceedings of the 2024 Conference on Empirical Methods in Natural Language Processing},
  year = {2024},
  pages = {1419--1436},
  publisher = {Association for Computational Linguistics},
  doi = {10.18653/v1/2024.emnlp-main.84},
  url = {https://aclanthology.org/2024.emnlp-main.84/}
}

@inproceedings{azaria2023internal,
  title = {The Internal State of an {LLM} Knows When It{'}s Lying},
  author = {Azaria, Amos and
      Mitchell, Tom},
  booktitle = {Findings of the Association for Computational Linguistics: EMNLP 2023},
  year = {2023},
  pages = {967--976},
  publisher = {Association for Computational Linguistics},
  doi = {10.18653/v1/2023.findings-emnlp.68},
  url = {https://aclanthology.org/2023.findings-emnlp.68/}
}

@inproceedings{orgad2024llms,
  title = {{LLM}s Know More Than They Show: On the Intrinsic Representation of {LLM} Hallucinations},
  author = {Orgad, Hadas and
      Toker, Michael and
      Gekhman, Zorik and
      Reichart, Roi and
      Szpektor, Idan and
      Kotek, Hadas and
      Belinkov, Yonatan},
  booktitle = {International Conference on Learning Representations},
  year = {2025},
  pages = {66880--66913},
  url = {https://proceedings.iclr.cc/paper_files/paper/2025/hash/a712d461e57201efe35d429a6f1731c1-Abstract-Conference.html}
}

@inproceedings{wang2025trustworthy,
  title = {Trustworthy Medical Question Answering: An Evaluation-Centric Survey},
  author = {Wang, Yinuo and
      Wang, Baiyang and
      Mercer, Robert and
      Rudzicz, Frank and
      Singha Roy, Sudipta and
      Ren, Pengjie and
      Chen, Zhumin and
      Wang, Xindi},
  booktitle = {Proceedings of the 2025 Conference on Empirical Methods in Natural Language Processing},
  year = {2025},
  pages = {27489--27502},
  publisher = {Association for Computational Linguistics},
  doi = {10.18653/v1/2025.emnlp-main.1398},
  url = {https://aclanthology.org/2025.emnlp-main.1398/}
}

@inproceedings{dong2025large,
  title = {Large Language Model Agents in Finance: A Survey Bridging Research, Practice, and Real-World Deployment},
  author = {Dong, Yifei and
      Wu, Fengyi and
      Zhang, Kunlin and
      Dai, Yilong and
      Zhang, Sanjian and
      Ye, Wanghao and
      Chen, Sihan and
      Cheng, Zhi-Qi},
  booktitle = {Findings of the Association for Computational Linguistics: EMNLP 2025},
  year = {2025},
  pages = {17889--17907},
  publisher = {Association for Computational Linguistics},
  doi = {10.18653/v1/2025.findings-emnlp.972},
  url = {https://aclanthology.org/2025.findings-emnlp.972/}
}

@article{wang2024survey,
  title = {A survey on large language model based autonomous agents},
  author = {Wang, Lei and
      Ma, Chen and
      Feng, Xueyang and
      Zhang, Zeyu and
      Yang, Hao and
      Zhang, Jingsen and
      Chen, Zhiyuan and
      Tang, Jiakai and
      Chen, Xu and
      Lin, Yankai and
      Zhao, Wayne Xin and
      Wei, Zhewei and
      Wen, Jirong},
  journal = {Frontiers of Computer Science},
  year = {2024},
  volume = {18},
  number = {6},
  pages = {186345},
  publisher = {Springer},
  doi = {10.1007/s11704-024-40231-1},
  url = {https://doi.org/10.1007/s11704-024-40231-1}
}

@article{reddy2019coqa,
  title = {{C}o{QA}: A Conversational Question Answering Challenge},
  author = {Reddy, Siva and
      Chen, Danqi and
      Manning, Christopher D.},
  journal = {Transactions of the Association for Computational Linguistics},
  year = {2019},
  volume = {7},
  pages = {249--266},
  publisher = {MIT Press},
  doi = {10.1162/tacl_a_00266},
  url = {https://aclanthology.org/Q19-1016/}
}

@inproceedings{rajpurkar2018squad,
  title = {Know What You Don{'}t Know: Unanswerable Questions for {SQ}u{AD}},
  author = {Rajpurkar, Pranav and
      Jia, Robin and
      Liang, Percy},
  editor = {Gurevych, Iryna and Miyao, Yusuke},
  booktitle = {Proceedings of the 56th Annual Meeting of the Association for Computational Linguistics (Volume 2: Short Papers)},
  month = jul,
  year = {2018},
  address = {Melbourne, Australia},
  pages = {784--789},
  publisher = {Association for Computational Linguistics},
  doi = {10.18653/v1/P18-2124},
  url = {https://aclanthology.org/P18-2124/}
}

@inproceedings{joshi2017triviaqa,
  title = {{T}rivia{QA}: A Large Scale Distantly Supervised Challenge Dataset for Reading Comprehension},
  author = {Joshi, Mandar and
      Choi, Eunsol and
      Weld, Daniel and
      Zettlemoyer, Luke},
  booktitle = {Proceedings of the 55th Annual Meeting of the Association for Computational Linguistics (Volume 1: Long Papers)},
  year = {2017},
  pages = {1601--1611},
  publisher = {Association for Computational Linguistics},
  doi = {10.18653/v1/P17-1147},
  url = {https://aclanthology.org/P17-1147/}
}

@article{kwiatkowski2019natural,
  title = {{Natural Questions}: A Benchmark for Question Answering Research},
  author = {Kwiatkowski, Tom and
      Palomaki, Jennimaria and
      Redfield, Olivia and
      Collins, Michael and
      Parikh, Ankur and
      Alberti, Chris and
      Epstein, Danielle and
      Polosukhin, Illia and
      Devlin, Jacob and
      Lee, Kenton and
      Toutanova, Kristina and
      Jones, Llion and
      Kelcey, Matthew and
      Chang, Ming-Wei and
      Dai, Andrew M. and
      Uszkoreit, Jakob and
      Le, Quoc and
      Petrov, Slav},
  journal = {Transactions of the Association for Computational Linguistics},
  year = {2019},
  volume = {7},
  pages = {452--466},
  publisher = {MIT Press},
  doi = {10.1162/tacl_a_00276},
  url = {https://aclanthology.org/Q19-1026/}
}

@misc{touvron2023llama,
  title = {{Llama 2}: Open Foundation and Fine-Tuned Chat Models},
  author = {Hugo Touvron and
      Louis Martin and
      Kevin Stone and
      Peter Albert and
      Amjad Almahairi and
      Yasmine Babaei and
      Nikolay Bashlykov and
      Soumya Batra and
      Prajjwal Bhargava and
      Shruti Bhosale and
      Dan Bikel and
      Lukas Blecher and
      Cristian Canton Ferrer and
      Moya Chen and
      Guillem Cucurull and
      David Esiobu and
      Jude Fernandes and
      Jeremy Fu and
      Wenyin Fu and
      Brian Fuller and
      Cynthia Gao and
      Vedanuj Goswami and
      Naman Goyal and
      Anthony Hartshorn and
      Saghar Hosseini and
      Rui Hou and
      Hakan Inan and
      Marcin Kardas and
      Viktor Kerkez and
      Madian Khabsa and
      Isabel Kloumann and
      Artem Korenev and
      Punit Singh Koura and
      Marie-Anne Lachaux and
      Thibaut Lavril and
      Jenya Lee and
      Diana Liskovich and
      Yinghai Lu and
      Yuning Mao and
      Xavier Martinet and
      Todor Mihaylov and
      Pushkar Mishra and
      Igor Molybog and
      Yixin Nie and
      Andrew Poulton and
      Jeremy Reizenstein and
      Rashi Rungta and
      Kalyan Saladi and
      Alan Schelten and
      Ruan Silva and
      Eric Michael Smith and
      Ranjan Subramanian and
      Xiaoqing Ellen Tan and
      Binh Tang and
      Ross Taylor and
      Adina Williams and
      Jian Xiang Kuan and
      Puxin Xu and
      Zheng Yan and
      Iliyan Zarov and
      Yuchen Zhang and
      Angela Fan and
      Melanie Kambadur and
      Sharan Narang and
      Aurelien Rodriguez and
      Robert Stojnic and
      Sergey Edunov and
      Thomas Scialom},
  year = {2023},
  doi = {10.48550/arXiv.2307.09288},
  eprint = {2307.09288},
  archivePrefix = {arXiv},
  primaryClass = {cs.CL},
  url = {https://arxiv.org/abs/2307.09288}
}

@inproceedings{ren2022out,
  title = {Out-of-distribution detection and selective generation for conditional language models},
  author = {Ren, Jie and
      Luo, Jiaming and
      Zhao, Yao and
      Krishna, Kundan and
      Saleh, Mohammad and
      Lakshminarayanan, Balaji and
      Liu, Peter J.},
  booktitle = {International Conference on Learning Representations},
  year = {2023},
  url = {https://iclr.cc/virtual/2023/poster/11478}
}

@inproceedings{liu2020energy,
  title = {Energy-based Out-of-distribution Detection},
  author = {Liu, Weitang and
      Wang, Xiaoyun and
      Owens, John and
      Li, Yixuan},
  booktitle = {Advances in Neural Information Processing Systems},
  year = {2020},
  volume = {33},
  pages = {21464--21475},
  publisher = {Curran Associates, Inc.},
  url = {https://proceedings.neurips.cc/paper_files/paper/2020/file/f5496252609c43eb8a3d147ab9b9c006-Paper.pdf}
}

@inproceedings{malinin2020uncertainty,
  title = {Uncertainty estimation in autoregressive structured prediction},
  author = {Malinin, Andrey and
      Gales, Mark},
  booktitle = {International Conference on Learning Representations},
  year = {2021},
  url = {https://iclr.cc/virtual/2021/poster/3015}
}

@inproceedings{liu2025attention,
  title = {Attention-guided Self-reflection for Zero-shot Hallucination Detection in Large Language Models},
  author = {Liu, Qiang and
      Chen, Xinlong and
      Ding, Yue and
      Song, Bowen and
      Wang, Weiqiang and
      Wu, Shu and
      Wang, Liang},
  booktitle = {Proceedings of the 2025 Conference on Empirical Methods in Natural Language Processing},
  year = {2025},
  pages = {21005--21021},
  publisher = {Association for Computational Linguistics},
  doi = {10.18653/v1/2025.emnlp-main.1063},
  url = {https://aclanthology.org/2025.emnlp-main.1063/}
}

@inproceedings{lin2004rouge,
  title = {{ROUGE}: A Package for Automatic Evaluation of Summaries},
  author = {Lin, Chin-Yew},
  booktitle = {Text Summarization Branches Out},
  year = {2004},
  pages = {74--81},
  publisher = {Association for Computational Linguistics},
  url = {https://aclanthology.org/W04-1013/}
}

@article{farquhar2024semantic,
  title = {Detecting hallucinations in large language models using semantic entropy},
  author = {Farquhar, Sebastian and
      Kossen, Jannik and
      Kuhn, Lorenz and
      Gal, Yarin},
  journal = {Nature},
  year = {2024},
  volume = {630},
  number = {8017},
  pages = {625--630},
  doi = {10.1038/s41586-024-07421-0},
  url = {https://doi.org/10.1038/s41586-024-07421-0}
}

@inproceedings{nikitin2024kernel,
  title = {Kernel Language Entropy: Fine-grained Uncertainty Quantification for {LLM}s from Semantic Similarities},
  author = {Nikitin, Alexander and
      Kossen, Jannik and
      Gal, Yarin and
      Marttinen, Pekka},
  booktitle = {Advances in Neural Information Processing Systems},
  year = {2024},
  volume = {37},
  pages = {8901--8929},
  publisher = {Curran Associates, Inc.},
  doi = {10.52202/079017-0283},
  url = {https://proceedings.neurips.cc/paper_files/paper/2024/file/10c456d2160517581a234dfde15a7505-Paper-Conference.pdf}
}

@inproceedings{li2025semanticgraph,
  title = {Enhancing Uncertainty Quantification in Large Language Models through Semantic Graph Density},
  author = {Li, Zhaoye and
      Shen, Siyuan and
      Yang, Wenjing and
      Jin, Ruochun and
      Chen, Huan and
      Cao, Ligong and
      Ren, Jing},
  booktitle = {Proceedings of the Forty-first Conference on Uncertainty in Artificial Intelligence},
  year = {2025},
  volume = {286},
  pages = {2537--2551},
  series = {Proceedings of Machine Learning Research},
  publisher = {PMLR},
  url = {https://proceedings.mlr.press/v286/li25b.html}
}

@inproceedings{kunitomo2026evidential,
  title = {Evidential Semantic Entropy for {LLM} Uncertainty Quantification},
  author = {Kunitomo-Jacquin, Lucie and
      Marrese-Taylor, Edison and
      Fukuda, Ken and
      Hamasaki, Masahiro},
  booktitle = {Proceedings of the 19th Conference of the {E}uropean Chapter of the {A}ssociation for {C}omputational {L}inguistics (Volume 1: Long Papers)},
  year = {2026},
  pages = {7107--7122},
  publisher = {Association for Computational Linguistics},
  doi = {10.18653/v1/2026.eacl-long.334},
  url = {https://aclanthology.org/2026.eacl-long.334/}
}

@misc{gemmateam2024gemma2,
  title = {{Gemma 2}: Improving Open Language Models at a Practical Size},
  author = {{Gemma Team} and
      Morgane Riviere and
      Shreya Pathak and
      Pier Giuseppe Sessa and
      Cassidy Hardin and
      Surya Bhupatiraju and
      Léonard Hussenot and
      Thomas Mesnard and
      Bobak Shahriari and
      Alexandre Ramé and
      Johan Ferret and
      Peter Liu and
      Pouya Tafti and
      Abe Friesen and
      Michelle Casbon and
      Sabela Ramos and
      Ravin Kumar and
      Charline Le Lan and
      Sammy Jerome and
      Anton Tsitsulin and
      Nino Vieillard and
      Piotr Stanczyk and
      Sertan Girgin and
      Nikola Momchev and
      Matt Hoffman and
      Shantanu Thakoor and
      Jean-Bastien Grill and
      Behnam Neyshabur and
      Olivier Bachem and
      Alanna Walton and
      Aliaksei Severyn and
      Alicia Parrish and
      Aliya Ahmad and
      Allen Hutchison and
      Alvin Abdagic and
      Amanda Carl and
      Amy Shen and
      Andy Brock and
      Andy Coenen and
      Anthony Laforge and
      Antonia Paterson and
      Ben Bastian and
      Bilal Piot and
      Bo Wu and
      Brandon Royal and
      Charlie Chen and
      Chintu Kumar and
      Chris Perry and
      Chris Welty and
      Christopher A. Choquette-Choo and
      Danila Sinopalnikov and
      David Weinberger and
      Dimple Vijaykumar and
      Dominika Rogozińska and
      Dustin Herbison and
      Elisa Bandy and
      Emma Wang and
      Eric Noland and
      Erica Moreira and
      Evan Senter and
      Evgenii Eltyshev and
      Francesco Visin and
      Gabriel Rasskin and
      Gary Wei and
      Glenn Cameron and
      Gus Martins and
      Hadi Hashemi and
      Hanna Klimczak-Plucińska and
      Harleen Batra and
      Harsh Dhand and
      Ivan Nardini and
      Jacinda Mein and
      Jack Zhou and
      James Svensson and
      Jeff Stanway and
      Jetha Chan and
      Jin Peng Zhou and
      Joana Carrasqueira and
      Joana Iljazi and
      Jocelyn Becker and
      Joe Fernandez and
      Joost van Amersfoort and
      Josh Gordon and
      Josh Lipschultz and
      Josh Newlan and
      Ju-yeong Ji and
      Kareem Mohamed and
      Kartikeya Badola and
      Kat Black and
      Katie Millican and
      Keelin McDonell and
      Kelvin Nguyen and
      Kiranbir Sodhia and
      Kish Greene and
      Lars Lowe Sjoesund and
      Lauren Usui and
      Laurent Sifre and
      Lena Heuermann and
      Leticia Lago and
      Lilly McNealus and
      Livio Baldini Soares and
      Logan Kilpatrick and
      Lucas Dixon and
      Luciano Martins and
      Machel Reid and
      Manvinder Singh and
      Mark Iverson and
      Martin Görner and
      Mat Velloso and
      Mateo Wirth and
      Matt Davidow and
      Matt Miller and
      Matthew Rahtz and
      Matthew Watson and
      Meg Risdal and
      Mehran Kazemi and
      Michael Moynihan and
      Ming Zhang and
      Minsuk Kahng and
      Minwoo Park and
      Mofi Rahman and
      Mohit Khatwani and
      Natalie Dao and
      Nenshad Bardoliwalla and
      Nesh Devanathan and
      Neta Dumai and
      Nilay Chauhan and
      Oscar Wahltinez and
      Pankil Botarda and
      Parker Barnes and
      Paul Barham and
      Paul Michel and
      Pengchong Jin and
      Petko Georgiev and
      Phil Culliton and
      Pradeep Kuppala and
      Ramona Comanescu and
      Ramona Merhej and
      Reena Jana and
      Reza Ardeshir Rokni and
      Rishabh Agarwal and
      Ryan Mullins and
      Samaneh Saadat and
      Sara Mc Carthy and
      Sarah Cogan and
      Sarah Perrin and
      Sébastien M. R. Arnold and
      Sebastian Krause and
      Shengyang Dai and
      Shruti Garg and
      Shruti Sheth and
      Sue Ronstrom and
      Susan Chan and
      Timothy Jordan and
      Ting Yu and
      Tom Eccles and
      Tom Hennigan and
      Tomas Kocisky and
      Tulsee Doshi and
      Vihan Jain and
      Vikas Yadav and
      Vilobh Meshram and
      Vishal Dharmadhikari and
      Warren Barkley and
      Wei Wei and
      Wenming Ye and
      Woohyun Han and
      Woosuk Kwon and
      Xiang Xu and
      Zhe Shen and
      Zhitao Gong and
      Zichuan Wei and
      Victor Cotruta and
      Phoebe Kirk and
      Anand Rao and
      Minh Giang and
      Ludovic Peran and
      Tris Warkentin and
      Eli Collins and
      Joelle Barral and
      Zoubin Ghahramani and
      Raia Hadsell and
      D. Sculley and
      Jeanine Banks and
      Anca Dragan and
      Slav Petrov and
      Oriol Vinyals and
      Jeff Dean and
      Demis Hassabis and
      Koray Kavukcuoglu and
      Clement Farabet and
      Elena Buchatskaya and
      Sebastian Borgeaud and
      Noah Fiedel and
      Armand Joulin and
      Kathleen Kenealy and
      Robert Dadashi and
      Alek Andreev},
  year = {2024},
  eprint = {2408.00118},
  archivePrefix = {arXiv},
  primaryClass = {cs.CL},
  url = {https://arxiv.org/abs/2408.00118}
}

@misc{qwen2024qwen25,
  title = {{Qwen2.5} Technical Report},
  author = {An Yang and
      Baosong Yang and
      Beichen Zhang and
      Binyuan Hui and
      Bo Zheng and
      Bowen Yu and
      Chengyuan Li and
      Dayiheng Liu and
      Fei Huang and
      Haoran Wei and
      Huan Lin and
      Jian Yang and
      Jianhong Tu and
      Jianwei Zhang and
      Jianxin Yang and
      Jiaxi Yang and
      Jingren Zhou and
      Junyang Lin and
      Kai Dang and
      Keming Lu and
      Keqin Bao and
      Kexin Yang and
      Le Yu and
      Mei Li and
      Mingfeng Xue and
      Pei Zhang and
      Qin Zhu and
      Rui Men and
      Runji Lin and
      Tianhao Li and
      Tianyi Tang and
      Tingyu Xia and
      Xingzhang Ren and
      Xuancheng Ren and
      Yang Fan and
      Yang Su and
      Yichang Zhang and
      Yu Wan and
      Yuqiong Liu and
      Zeyu Cui and
      Zhenru Zhang and
      Zihan Qiu},
  year = {2024},
  eprint = {2412.15115},
  archivePrefix = {arXiv},
  primaryClass = {cs.CL},
  url = {https://arxiv.org/abs/2412.15115}
}

@misc{jiang2023mistral,
  title = {{Mistral 7B}},
  author = {Albert Q. Jiang and
      Alexandre Sablayrolles and
      Arthur Mensch and
      Chris Bamford and
      Devendra Singh Chaplot and
      Diego de las Casas and
      Florian Bressand and
      Gianna Lengyel and
      Guillaume Lample and
      Lucile Saulnier and
      Lélio Renard Lavaud and
      Marie-Anne Lachaux and
      Pierre Stock and
      Teven Le Scao and
      Thibaut Lavril and
      Thomas Wang and
      Timothée Lacroix and
      William El Sayed},
  year = {2023},
  eprint = {2310.06825},
  archivePrefix = {arXiv},
  primaryClass = {cs.CL},
  url = {https://arxiv.org/abs/2310.06825}
}

@inproceedings{yang2018hotpotqa,
  title = {{H}otpot{QA}: A Dataset for Diverse, Explainable Multi-hop Question Answering},
  author = {Yang, Zhilin and
      Qi, Peng and
      Zhang, Saizheng and
      Bengio, Yoshua and
      Cohen, William and
      Salakhutdinov, Ruslan and
      Manning, Christopher D.},
  booktitle = {Proceedings of the 2018 Conference on Empirical Methods in Natural Language Processing},
  year = {2018},
  pages = {2369--2380},
  publisher = {Association for Computational Linguistics},
  doi = {10.18653/v1/D18-1259},
  url = {https://aclanthology.org/D18-1259/}
}

@inproceedings{niu2024ragtruth,
  title = {{RAGT}ruth: A Hallucination Corpus for Developing Trustworthy Retrieval-Augmented Language Models},
  author = {Niu, Cheng and
      Wu, Yuanhao and
      Zhu, Juno and
      Xu, Siliang and
      Shum, KaShun and
      Zhong, Randy and
      Song, Juntong and
      Zhang, Tong},
  booktitle = {Proceedings of the 62nd Annual Meeting of the Association for Computational Linguistics (Volume 1: Long Papers)},
  year = {2024},
  pages = {10862--10878},
  publisher = {Association for Computational Linguistics},
  doi = {10.18653/v1/2024.acl-long.585},
  url = {https://aclanthology.org/2024.acl-long.585/}
}

@inproceedings{chen2023felm,
  title = {{FELM}: Benchmarking Factuality Evaluation of Large Language Models},
  author = {Chen, Shiqi and
      Zhao, Yiran and
      Zhang, Jinghan and
      Chern, I-Chun and
      Gao, Siyang and
      Liu, Pengfei and
      He, Junxian},
  booktitle = {Advances in Neural Information Processing Systems},
  year = {2023},
  volume = {36},
  pages = {44502--44523},
  publisher = {Curran Associates, Inc.},
  doi = {10.52202/075280-1927},
  url = {https://proceedings.neurips.cc/paper_files/paper/2023/file/8b8a7960d343e023a6a0afe37eee6022-Paper-Datasets_and_Benchmarks.pdf}
}

@inproceedings{bakman2024mars,
  title = {{MARS}: Meaning-Aware Response Scoring for Uncertainty Estimation in Generative {LLM}s},
  author = {Bakman, Yavuz Faruk and
      Yaldiz, Duygu Nur and
      Buyukates, Baturalp and
      Tao, Chenyang and
      Dimitriadis, Dimitrios and
      Avestimehr, Salman},
  booktitle = {Proceedings of the 62nd Annual Meeting of the Association for Computational Linguistics (Volume 1: Long Papers)},
  year = {2024},
  pages = {7752--7767},
  publisher = {Association for Computational Linguistics},
  doi = {10.18653/v1/2024.acl-long.419},
  url = {https://aclanthology.org/2024.acl-long.419/}
}

@inproceedings{aichberger2025sdlg,
  title = {Improving Uncertainty Estimation through Semantically Diverse Language Generation},
  author = {Aichberger, Lukas and
      Schweighofer, Kajetan and
      Ielanskyi, Mykyta and
      Hochreiter, Sepp},
  booktitle = {International Conference on Learning Representations},
  year = {2025},
  pages = {74479--74503},
  url = {https://proceedings.iclr.cc/paper_files/paper/2025/file/b94d8b035e2183e47afef9e2f299ba47-Paper-Conference.pdf}
}


%% file: arxiv_extra.bib
@article{zhang2026conffuse,
  title = {Global-Local Confidence Fusion for Hallucination Detection in Mathematical Reasoning Task},
  author = {Zhang, Bo and Gao, Cong and Yang, Linkang and Han, Bingxu and Hu, Minghao and Luo, Zhunchen and Geng, Guotong and Bai, Xiaoying and Zhang, Jun and Yao, Wen and Wang, Zhong},
  journal = {Proceedings of the AAAI Conference on Artificial Intelligence},
  volume = {40},
  number = {41},
  pages = {34620--34628},
  year = {2026},
  doi = {10.1609/aaai.v40i41.40762},
  url = {https://ojs.aaai.org/index.php/AAAI/article/view/40762}
}

@inproceedings{corpataux2026whisper,
  title = {Detecting {Whisper} Hallucinations with Local Confidence Contrasts},
  author = {Corpataux, Sam and Scius-Bertrand, Anna and Wolf, Beat},
  booktitle = {Proceedings of the Fourth Swiss AI Days},
  series = {Proceedings of Machine Learning Research},
  volume = {309},
  pages = {38--45},
  year = {2026},
  publisher = {PMLR},
  url = {https://proceedings.mlr.press/v309/corpataux26a.html}
}

@article{kiss-strunk-2006-unsupervised,
  title = {Unsupervised Multilingual Sentence Boundary Detection},
  author = {Kiss, Tibor and Strunk, Jan},
  journal = {Computational Linguistics},
  volume = {32},
  number = {4},
  pages = {485--525},
  year = {2006},
  doi = {10.1162/coli.2006.32.4.485},
  url = {https://aclanthology.org/J06-4003/}
}


%% file: arxiv_multilingual.bib
@inproceedings{vazquez2025mushroom,
  title = {{SemEval}-2025 Task 3: {Mu-SHROOM}, the Multilingual Shared-task on Hallucinations and Related Observable Overgeneration Mistakes},
  author = {Vazquez, Raul and Mickus, Timothee and Zosa, Elaine and Vahtola, Teemu and Tiedemann, J{\"o}rg and Sinha, Aman and Segonne, Vincent and Sanchez-Vega, Fernando and Raganato, Alessandro and Libovick{\'y}, Jind{\v{r}}ich and Karlgren, Jussi and Ji, Shaoxiong and Helcl, Jind{\v{r}}ich and Guillou, Liane and De Gibert, Ona and Bengoetxea, Jaione and Attieh, Joseph and Apidianaki, Marianna},
  booktitle = {Proceedings of the 19th International Workshop on Semantic Evaluation (SemEval-2025)},
  year = {2025},
  pages = {2472--2497},
  publisher = {Association for Computational Linguistics},
  url = {https://aclanthology.org/2025.semeval-1.322/}
}

@inproceedings{kamei2025jhars,
  title = {{JHARS}: Construction and Analysis of a Japanese Hallucination Evaluation Benchmark in {RAG} Settings},
  author = {Kamei, Ryohei and Sakata, Masaki and Hentona, Asahi and Kurihara, Kentaro and Inui, Kentaro},
  booktitle = {Proceedings of the Thirty-first Annual Meeting of the Association for Natural Language Processing},
  year = {2025},
  pages = {833--838},
  note = {In Japanese},
  url = {https://www.anlp.jp/proceedings/annual_meeting/2025/pdf_dir/Q2-17.pdf}
}

@inproceedings{chatterjee2026hat,
  title = {{HAT}: Hallucination Annotation for Translation},
  author = {Chatterjee, Rajen and Li, Xintong and Charoenpornsawat, Paisarn and Lee, Allen},
  booktitle = {Proceedings of the 64th Annual Meeting of the Association for Computational Linguistics (Volume 1: Long Papers)},
  year = {2026},
  pages = {15865--15888},
  publisher = {Association for Computational Linguistics},
  doi = {10.18653/v1/2026.acl-long.721},
  url = {https://aclanthology.org/2026.acl-long.721/}
}
